\documentclass{ieeeaccess}
\IEEEoverridecommandlockouts
\usepackage{cite}
\usepackage{amsmath,amssymb,amsfonts}
\usepackage{algorithmic}
\usepackage{graphicx}
\usepackage{textcomp}
\usepackage{bm}
\usepackage{multirow}
\usepackage{makecell}
\usepackage{siunitx}
\newcommand{\rev}[1]{\textcolor{black}{#1}}

\makeatletter
\AtBeginDocument{\DeclareMathVersion{bold}
\SetSymbolFont{operators}{bold}{T1}{times}{b}{n}
\SetSymbolFont{NewLetters}{bold}{T1}{times}{b}{it}
\SetMathAlphabet{\mathrm}{bold}{T1}{times}{b}{n}
\SetMathAlphabet{\mathit}{bold}{T1}{times}{b}{it}
\SetMathAlphabet{\mathbf}{bold}{T1}{times}{b}{n}
\SetMathAlphabet{\mathtt}{bold}{OT1}{pcr}{b}{n}
\SetSymbolFont{symbols}{bold}{OMS}{cmsy}{b}{n}
\renewcommand\boldmath{\@nomath\boldmath\mathversion{bold}}}
\def\theHequation{\theequation}
\makeatother

\def\BibTeX{{\rm B\kern-.05em{\sc i\kern-.025em b}\kern-.08em
    T\kern-.1667em\lower.7ex\hbox{E}\kern-.125emX}}
    
\usepackage{array}
\newcolumntype{P}[1]{>{\raggedright\arraybackslash}p{#1}}
\newcolumntype{C}[1]{>{\centering\arraybackslash}p{#1}}

\ifCLASSOPTIONcompsoc
\usepackage[caption=false,font=normalsize,labelfont=sf,textfont=sf]{subfig}
\else
\usepackage[caption=false,font=footnotesize]{subfig}
\fi

\usepackage[hidelinks]{hyperref}

\begin{document}
\history{Date of publication xxxx 00, 0000, date of current version xxxx 00, 0000.}
\doi{xx}
\title{Integration of UWB Radar on Mobile Robots for Continuous Obstacle and Environment Mapping}

\author{\uppercase{Adelina Giurea}\authorrefmark{1},
\uppercase{Stijn Luchie}\authorrefmark{1}, \uppercase{Dieter Coppens}\authorrefmark{1}, \uppercase{Jeroen Hoebeke}\authorrefmark{1}, 
\uppercase{Eli De Poorter}\authorrefmark{1}}

\address[1]{Department of Information Technology, Ghent University - imec - IDLab, Technologiepark 126, 9052 Ghent, Belgium}
\tfootnote{The research that led to these results was partly funded by the DistriMuSe project (HORIZON-KDT-JU-2023-2-RIA HORIZON JU Research and Innovation Actions) with Grant No 101139769}

\markboth
{A. Giurea \headeretal: Integration of UWB Radar on Mobile Robots for Continuous Obstacle and Environment Mapping}
{A. Giurea \headeretal: Integration of UWB Radar on Mobile Robots for Continuous Obstacle and Environment Mapping}

\corresp{Corresponding author: Adelina Giurea (e-mail: adelina.giurea@ugent.be)}

\begin{abstract}
This paper presents an infrastructure-free approach for obstacle detection and environmental mapping using ultra-wideband (UWB) radar mounted on a mobile robotic platform. Traditional sensing modalities such as visual cameras and Light Detection and Ranging (LiDAR) fail in environments with poor visibility due to darkness, smoke, or reflective surfaces. \rev{In these vision-impaired conditions, UWB radar offers a promising alternative. To this end, this work explores the suitability of robot-mounted UWB radar for environmental mapping in anchor-free, unknown scenarios.} The study investigates how different materials (metal, concrete and plywood) and UWB radio channels (5 and 9) influence the Channel Impulse Response (CIR). Furthermore, a processing pipeline is proposed to achieve reliable mapping of detected obstacles, consisting of 3 steps: (i) target identification (based on CIR peak detection), (ii) filtering (based on peak properties, signal-to-noise score, and phase-difference of arrival), and (iii) clustering (based on distance estimation and angle-of-arrival estimation). \rev{The proposed approach successfully reduces noise and multipath effects, achieving high obstacle detection performance across a range of materials. Even in challenging low-reflectivity scenarios such as concrete, the method achieves a precision of $73.42\%$ and a recall of $83.38\%$ on channel 9.} This work offers a foundation for further development of UWB-based localisation and mapping (SLAM) systems that do not rely on visual features and, unlike conventional UWB localisation systems, do not require fixed anchor nodes for triangulation.

\end{abstract}


\titlepgskip=-21pt

\maketitle

\section{Introduction}
\label{sec:introduction}
\PARstart{T}{he} ability to detect objects and build a representation of the surrounding environment is an essential requirement to allow autonomous operations of mobile robots. Existing systems rely on cameras to capture images and track movement by extracting visual features. However, this method depends on clear lighting conditions and struggles in environments with low visibility, shadows, or reflective surfaces \cite{review-LiDAR}. Light Detection and Ranging (LiDAR) systems, on the other hand, use laser pulses to measure distances and construct maps. While highly accurate, LiDAR sensors face challenges in environments where laser beams are scattered or absorbed, such as in heavy rain, fog, or areas with multiple reflective surfaces. 
Given these challenges, there is a growing need for alternative localisation and mapping technologies. Ultra-wideband (UWB) is a promising alternative due to its ability to operate in low-visibility environments while maintaining high accuracy (typically under 20 cm). However, UWB-based localisation systems usually rely on anchor-tag setups, where fixed UWB anchors are installed in predefined locations, and mobile robots are equipped with UWB-tags \cite{ensemble-learning} \cite{nlos-neural-network}. Although this setup works well for environments with stable infrastructure, it is impractical in unknown places, such as search and rescue activities or industrial facilities where the surroundings frequently change. \rev{Furthermore, using a UWB radio instead of LiDAR can reduce the system’s total weight, a significant advantage for drones to maximize their autonomy \cite{survey-robot-swarms}.}

To realize this vision, this paper proposes and evaluates signal processing methods to detect obstacles and map the environment. For evaluating the proposed algorithm, data were collected using the Qorvo QM33120WDK1 development kit \cite{datasheet-sensors}. The setup consists of separate omnidirectional transmitter antenna and directional receiver antennas mounted on a TurtleBot 4.
The main contributions of this paper are as follows:
\begin{itemize}
    \item We analyse how three common obstacle materials (metal, concrete, and plywood) affect the Channel Impulse Response (CIR), and identify optimal signal parameters (peak width and prominence) for obstacle detection of these materials across IEEE 802.15.4 UWB channels 5 and 9.
    \item We propose a novel noise filtering method that eliminates multipath-induced phantom reflections by computing a reliability score based on peak characteristics, signal-to-noise ratio (SNR) and phase-difference of arrival (PDoA).
    \item  \rev{We improve UWB obstacle detection by combining angle-of-arrival (AoA) and distance to cluster radar returns consistently across time, producing stable and spatially coherent obstacle maps.}
    \item We evaluate the complete processing pipeline in realistic conditions within a large industrial testbed environment.
    \item We publicly release the captured dataset, processing pipeline, and an illustrative video to support reproducibility and further research\footnote{https://gitlab.ilabt.imec.be/datasets/uwb-radar-obstacle-mapping-dataset}.
\end{itemize}

\rev{The remainder of this paper is organised as follows: Section \ref{sec:background-uwb} provides background information on UWB. Section \ref{sec:related-work} discusses related work, whereas Section \ref{sec:proposed-approach} presents the proposed processing pipeline. In Section \ref{sec:data-collection}, the data collection is explained, and the results for the proposed method are presented in Section \ref{sec:results} together with a visual demonstration. Finally, Sections \ref{sec:discussion} and \ref{sec:conclusion} cover limitations and the conclusion with suggestions for future work, respectively.}

\section{Background of Ultra-Wideband}
\label{sec:background-uwb}

UWB is a radio technology that transmits data over a wide frequency range, either with an absolute bandwidth greater than 500 MHz or a fractional bandwidth (\(B_F\)) greater than 20\%. The fractional bandwidth (\(B_F\)) is defined as the ratio of the signal's bandwidth (\(B_W\)) to its centre frequency (\(f_C\)) as:

\[
B_F =\frac{BW}{f_C}=\frac{(f_H - f_L)}{(f_H + f_L)/2}
\]

Where \( f_H \) and \( f_L \) are the upper and lower frequencies of the -10 dB bandwidth \cite{uwb-overview}.

Because UWB technology uses a large bandwidth, it enables the transmission of extremely short, narrow pulses in the time domain as described by the time-bandwidth relation \( \text{BW} \times T \geq \frac{1}{\pi} \), where  (\(B_W\)) is the signal's bandwidth and (\(T\)) the pulse duration. For example, UWB systems using a 500 MHz bandwidth can generate pulses of only 0.16 ns wide, whereas traditional technologies such as Wi-Fi, which are limited to bandwidths around 20 MHz, produce pulses longer than 4 ns. Due to its high bandwidth and ultra-narrow pulses, UWB is considered to have a high range resolution and ability to distinguish multiple targets.

This technology operates within the frequency spectrum of 3.1 GHz to 10.6 GHz, enabling data transmission speeds of up to 110 Mbps, typically with a range of up to 10 meters. A higher range of up to 100 meters is also possible, but at lower data rates, depending on several factors, including channel frequency, antenna design, power levels, and complexity of the propagation environment \cite{uwb-standards}.

\rev{UWB radar can be implemented either as Direct Sequence UWB (DS-UWB), which spreads a carrier signal using pseudo-random noise code, or as Impulse Radio UWB (IR-UWB), which transmits short pulses and estimates distance using the Time-of-Flight (ToF). Although pseudo-random noise radars are common in scientific research, they are typically not compatible with IEEE 802.15.4 UWB localisation and communication systems, and no low-cost commercial radio chips exist. For this reason, this proposed approach focuses on pulse-based UWB.}

\begin{figure}
    \centering
    \includegraphics[width=1\linewidth]{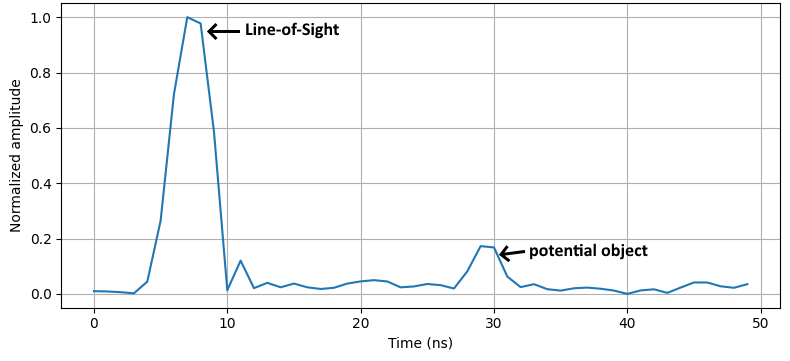}
    \caption{Example of a Channel Impulse Response where the object is placed at a distance of 3.5 m.}
    \label{fig:CIR}
\end{figure}

\rev{To better understand how IR-UWB radar works, it is essential to study how the transmitted signal interacts with obstacles and reflections in the environment. This is described by the CIR, illustrated in Fig. \ref{fig:CIR}. The first peak in the figure corresponds to the Line-of-Sight (LoS) component. In this case, the LoS signal indicates the signal transmitted from the transmitter antenna to the receiver antenna co-located on the same robot. Subsequent peaks represent reflections from other objects, walls, or multipath components.}

The CIR is typically modelled as a sum of deterministic and diffuse multipath components \cite{cir-formula} as shown in \eqref{eq:cir_superposition}:

\rev{\begin{equation}
    CIR(t) = \sum_{i} A_i \delta(t - \tau_i) + \nu(t) 
    \label{eq:cir_superposition}
\end{equation}}

\rev{The first term represents the deterministic multipath components, which describe the propagation of the UWB signal through multipaths with different delays. Each component is characterized by an amplitude \(A_i\) and a corresponding delay \(\tau_i\). The second term accounts for measurement noise, modelled as additive white Gaussian noise (AWGN), denoted by \(\nu(t)\).}

In UWB radar systems, the CIR data is typically obtained as I$/$Q samples (In-Phase and Quadrature components) of the received signal, which provides important information about the signal's strength and phase.

\rev{UWB radar is widely used in applications that require precise distance measurements, even in cluttered or unknown environments. Some examples of these applications are presence detection, device-free localisation, monitoring vital signs without physical contact \cite{lambrecht2025lowcostembeddedbreathingrate} and, activity recognition. UWB is a strong candidate for fast-moving robots and unknown environments due to its low weight, low power consumption, and cost-effectiveness \cite{survey-robot-swarms}.} While RGB cameras are also lightweight and energy-efficient, they suffer in complex or low-light environments. LiDAR offers high-accuracy long-range sensing but is heavy and power-intensive. LiDAR performance degrades in environments with smoke, dust, or fire, as laser beams cannot penetrate these particles. Additionally, both LiDAR and depth cameras have difficulties detecting transparent objects. Vanhie-Van Gerwen et al. \cite{indoor-drone} show that across various drone sizes, UWB stands out for its favourable accuracy-to-cost ratio. It also provides precise positioning thanks to its wide bandwidth and short pulses, is resistant to narrowband interference, and respects privacy by not capturing identity-related data. Both mmWave and UWB can distinguish small objects thanks to their high bandwidth. However, signals with lower centre frequencies generally propagate farther. mmWave operates between 30 GHz and 300 GHz, resulting in high-speed data transmission, but the technology suffers from high path loss, limited penetration, and short range. In contrast, UWB experiences less attenuation over distance and better penetration of walls and obstacles.

\section{Related Work}
\label{sec:related-work}

\begin{table*}[t]
    \caption{Overview of Related Work in UWB-Based Object Detection and Distance Estimation}
    \begin{center}
    \renewcommand{\arraystretch}{1.2}
    \resizebox{\textwidth}{!}{%
    \begin{tabular}{w{c}{0.9cm}|P{2.5cm}P{2.5cm}P{2cm}p{1.5cm}C{1.5cm}C{1.5cm}C{1.5cm}C{1.5cm}P{1.8cm}P{2.2cm}}
    \hline
    \textbf{Paper} & \textbf{Goal} & \textbf{Approach} & \textbf{Chip} & \textbf{UWB Mobile robot}& \textbf{No Anchor Nodes} & \textbf{Distance Estimation} & \textbf{Angle Estimation of Objects} & \textbf{IEEE 802.15.4 Compliant} & \textbf{Target Object} & \textbf{Detection Accuracy} \\
    \hline
    \hline
    \cite{cir-processing} & Object/person detection using CIR analysis & Three CIR processing techniques (ICIR, UCIR, ACIR) with BMA and BSDA algorithms & DW1000 & - & \checkmark & \checkmark  & -  & \checkmark & whiteboard, large tv, metal box, person & Avg. error $\leq$ 9 cm using ACIR (larger for persons or smaller objects)  \\
    \hline
    \cite{machine-learning-train} & Detect humans/obstacles in automatic train pairing. & Supervised Machine Learning  & P440 from TimeDomain & - & \checkmark & - & -  & - & humans/obstacles  & Best detection accuracy: 55\%-70\% indoors,$>$ 95\% outdoors  \\
    \hline
    \cite{distance-estimation-ir-uwb} & Remove clutter to improve detection accuracy & Low-rank approximation and singular value decomposition methods & not specified & - & \checkmark & \checkmark & - & - & walking person & Avg. error $< 20$ cm, RMSE $< 50$ cm \\
    \hline
    \cite{LSTM-based-system-multiple-obstacle-detection} & Distinguish objects from noise in transportation use cases. & Uses Discrete Wavelet Transform to extract features and forward these to the LSTM network. & UMAIN HST-D3 directional antenna & - & \checkmark & - & - & - & pedestrian, cyclist, vehicle, tram  & 72.78\% precision, 71.34\% recall, 72.06\% F1-score \\
    \hline
    \cite{entropy-based} & Detect the target object & Utilises signal's entropy  & UMAIN HST-D3 directional antenna & - & \checkmark & -  & - & - & pedestrian, cyclist, vehicle, tram &  75.34\% precision, 63.06\% recall, 68.65\% F1-score \\
    \hline

    \cite{localisation-navigation-uwb} & Signal processing chain for obstacle detection & Background subtraction, CFAR and Kalman filter & M-sequence UWB & \checkmark & \checkmark & \checkmark & - & - & wall and glass pane & not specified \\
    \hline
    \cite{cooperative-uwb-slam} & UWB-SLAM & Kalman Filter to reduce followed by particle filter & M-Sequence UWB & \checkmark & \checkmark & \checkmark & - & - &walls and corners & $\leq$ 20 cm\\
    \hline
    \cite{anchorless-uwb-1} & UWB-SLAM & wheel odometry and Extended Kalman Filter & X4M300 & \checkmark & \checkmark & \checkmark & - & - & vertical metal rods & 6.2 cm localisation error \\
    \hline
    \cite{anchorless-uwb-2} & UWB-SLAM in feature-deficient areas & Extends previous approach with UWB AoA anchor-tag for better loop closure& X4M300 & \checkmark & - &\checkmark & - & - & vertical metal rods & 10.3 cm localisation error \\
    \hline
    \hline
    This paper & Reduce noise and unwanted multipath for reliable obstacle mapping & Filtering on peak properties, SNR-score, PDoA and clustering & QM33120WDK1 & \checkmark & \checkmark &\checkmark & \checkmark & \checkmark & metal plate, concrete and plywood box & \rev{MAE $\leq$ 16.70 cm, $\geq 73.42 \%$ precision, $\geq 83.38 \%$ recall on channel 9 }\\
    \hline
    \hline
    \end{tabular}}
    \label{tab:related_work}
    \end{center}
\end{table*}
UWB is widely known for its use in localisation and anchor-tag positioning, where fixed UWB infrastructure (`UWB anchor nodes') is used to localise mobile robots. Although highly accurate, these localisation approaches require significant infrastructure investments and are hence outside the scope of this paper. Instead, in this section, we discuss prior works that focused on obstacle detection using fixed or mobile UWB radar systems and compare how they differ from our work. 

Table \ref{tab:related_work} shows a comparison of the existing work with our approach. Most state-of-the-art solutions rely on the installation of fixed UWB radar systems. 
For example, Van Herbruggen et al. \cite{cir-processing} introduce Accumulation Channel Impulse Response (ACIR) technique, which improves signal resolution for obstacle detection. Although this method increases the detection accuracy, it requires the collection of sufficient samples before it can be applied, limiting the maximum speed of robotic platforms. \rev{Sattiraju et al. \cite{machine-learning-train} conclude that Random Forest and Extra Trees achieve high accuracy in obstacle detection. However, such learning-based methods are prone to overfitting the training data, which can reduce generalisation to new environments. Yun et al. \cite{distance-estimation-ir-uwb} suppress noise from the received signal using singular value decomposition (SVD). Although their method achieves high accuracy, the required matrix computations are computationally intensive and may increase the latency and memory usage for real-time mobile robots. Mimouna et al. \cite{LSTM-based-system-multiple-obstacle-detection}, \cite{entropy-based} focus on obstacle detection, but lack characterisation of the obstacle (e.g. estimation of distance and width). Moreover, the use of complex deep learning models in \cite{LSTM-based-system-multiple-obstacle-detection} would lead to a higher execution time.}

\rev{In general, learning-based methods are well suited for capturing complex signal characteristics and can achieve high detection accuracy when trained on sufficiently large and representative datasets. However, their performance strongly depends on the quality and diversity of the training data and may degrade when deployed in environments that differ from those seen during training. As a result, these methods are more prone to overfitting and reduced generalisation in unseen environments.}

C. Smeenk et al. \cite{localisation-navigation-uwb} present a signal processing chain for M-sequence UWB radars mounted on a mobile robot. Their approach combines background subtraction to remove the antenna crosstalk, with a constant false alarm rate (CFAR) method for object detection and distance estimation. However, the localisation is limited to reporting that an object is in front of the antenna and its distance, without providing its relative position in the environment.
\rev{The method presented in this work overcomes all the above limitations by using clustering that is more adaptable to unseen environments. Moreover, our work also characterizes the position and width of detected obstacles by combining both distance estimation and AoA information.}

To the best of our knowledge, only four prior publications focus on robot-mounted UWB radar systems. \cite{anchorless-uwb-1} and \cite{anchorless-uwb-2} both focus more on Simultaneous Localisation and Mapping (SLAM) localisation aspects than UWB signal processing optimisations for obstacle characterisation. They are also limited by recognizing only vertical rods rather than reconstructing full obstacle representations from various material types. The proposed approach from our work can serve as a complementary method to \cite{anchorless-uwb-1} and \cite{anchorless-uwb-2}, aiming to suppress noise and mitigate ghost detections while increasing the overall performance of their proposed UWB-SLAM algorithm. Unlike \cite{anchorless-uwb-2}, which complements UWB-SLAM with localisation by employing AoA nodes for robot localisation and loop closure, our paper focuses on estimating the AoA of detected objects using the PDoA between antennas on the same chip. R. Zetik et al. \cite{cooperative-uwb-slam} also focus on SLAM and demonstrate good localisation accuracy across different room shapes. However, their approach relies on proprietary antenna designs tailored for their experiments, which increases deployment cost.

Finally, almost all prior work relies on custom UWB radar-only hardware devices with larger bandwidths and thus higher range resolution. Instead, our proposed approach is fully compliant with low-cost IEEE 802.15.4 standard off-the-shelf hardware. Using this setup, the same hardware can also localise the robot if fixed infrastructure is available, by reusing the transmitted wireless signal for multiple purposes simultaneously. This results in reduced energy consumption, reduced need for separate hardware, lower costs and weight, and more efficient use of the sparse spectrum. In this way, low-cost joint sensing, communication, and localisation for robotic platforms can be realised \cite{integrated-sensing-and-communication}.

\section{Proposed approach}
\label{sec:proposed-approach}

\begin{figure}
    \centering
    \includegraphics[width=1\linewidth]{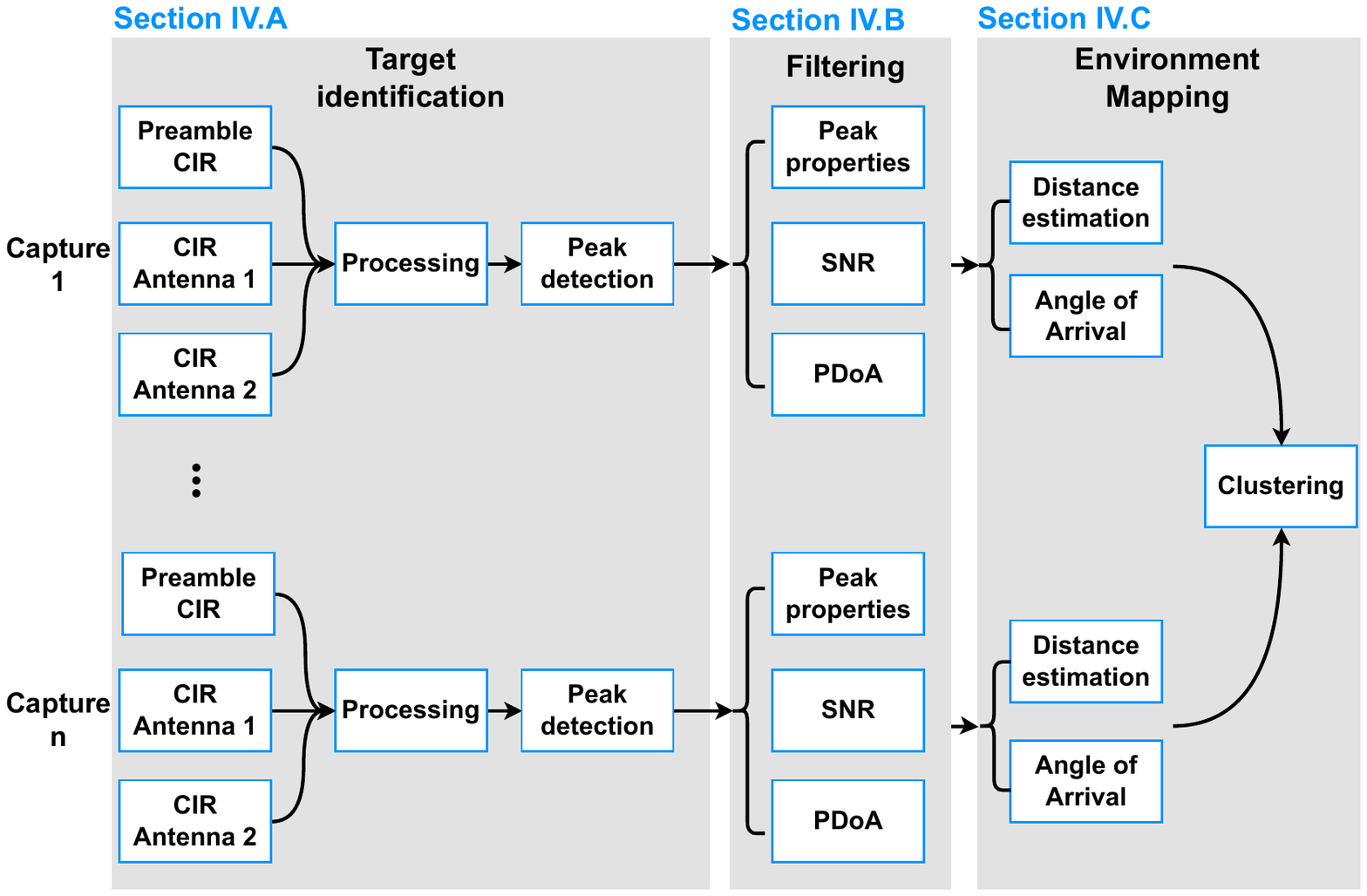}
    \caption{Overview of the proposed approach: Each capture contains the preamble CIR and two additional CIRs from both antennas on the same chip used for calculating the PDoA. A new measurement happens every 10.42 ms and contains the data of the environment at one timestamp. These are processed independently followed by filtering to remove noise and unwanted multipath components using peak properties, SNR-score, and PDoA. The distance to the object is estimated and the AoA is then calculated. Subsequently, multiple rows are combined to cluster reflections from the same object and map the environment.}
    \label{fig:stappenplan}
\end{figure}
This section outlines the complete workflow of the proposed approach for environment mapping. As illustrated in Fig. \ref{fig:stappenplan}, it consists of three main steps: (i) target identification, (ii) signal filtering, and (iii) the combination of the individual observations for obstacle and environment mapping.

\subsection{Step 1: Target Identification}
\label{subsec:obstacle-detection}
\begin{figure}
    \centering
    \includegraphics[width=1\linewidth]{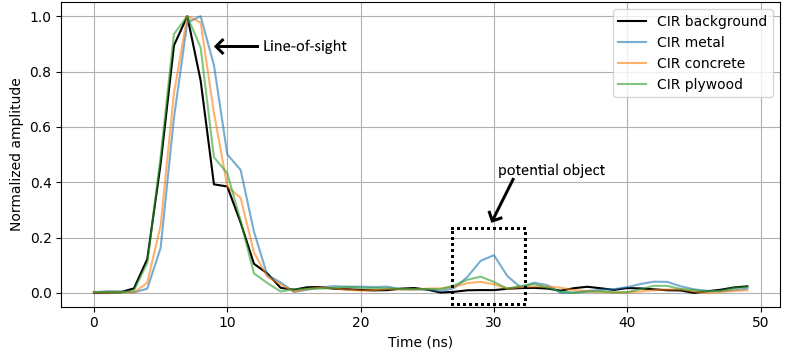}
    \caption{The CIRs measured for metal, concrete and plywood objects, overlaid on a background CIR measured without any obstacles. The reflections caused by each material appear as distinct peaks at specific delays relative to the first path. The metal object (blue) is clearly visible compared to the background.}
    \label{fig:CIRs}
\end{figure}

\rev{UWB's large bandwidth and narrow pulses give it high time resolution, allowing precise separation of multiple reflected signals. Fig. \ref{fig:CIRs} illustrates this by showing the CIRs measured for metal, concrete, and plywood obstacles, overlaid on a background CIR without obstacles. Each material produces distinct reflection peaks at specific delays relative to the first path.}

The UWB receiver generates three types of CIRs:
\begin{itemize}
    \item \textit{Preamble CIR}: captured on the first antenna and used for distance estimation.
    \item \textit{STS1-CIR}: obtained from the Scrambled Timestamp Sequence (STS) on the first antenna.
    \item \textit{STS2-CIR}: obtained from the STS after the chip switches to the second antenna.
\end{itemize}
The QM331x0W chips, which comply with the IEEE 802.15.4z standard, integrate two antennas on the same chip. This standard introduces the STS field mainly for security purposes, but it also enables the creation of additional CIRs that can be used for angle of arrival (AoA) estimation as the same signal reaches each antenna with slightly different phase. On the first antenna, the receiver captures both the preamble CIR and the STS1-based CIR. Then, halfway through the UWB packet, the chip switches to the second antenna to capture the STS2-based CIR \cite{single-anchor-localization}.

\begin{table}
    \caption{Mathematical symbols used throughout this paper}
        \begin{center}
        \renewcommand{\arraystretch}{1.2}
        \begin{tabular}{rP{4.5cm}P{1cm}}
        \hline
        \textbf{Variable} & \textbf{Description} & \textbf{Unit} \\
        \hline
        \hline
        \textit{\(B_F\)} & Fractional bandwidth & - \\
        
        \textit{\(B_W\)} & Signal's bandwidth & Hz\\
        
        \textit{\(F_C\)} & Centre frequency & Hz \\
        
        \textit{\(F_H\) and \(F_L\)} & Upper and lower frequency of the -10 dB bandwidth & Hz \\
        
        \textit{\(CIR(t)\)} & Channel Impulse Response & - \\
        
        \textit{\(CIR_{magnitude}\)}& Magnitude of a complex I/Q sample & - \\
        
        \textit{\(\phi\)} & Phase of an I/Q sample & radians \\
        
        $\mathit{SNR\text{-}score}_{i}$ & SNR-based score for peak with index \textit{i} & - \\
    
        \textit{\(A_i\)} & Amplitude of peak with index \textit{i} in the CIR & - \\
        
        \textit{\(A_{noise}\)} & The RMS of noise floor in the CIR & - \\
        
        \textit{\(k\)} & Linear weighting factor for calculating the SNR-score & - \\

        \textit{\(d_{TX-p}\)} & Distance between transmitter and object & cm \\
        
        \textit{\(d_{RX-p}\)} & Distance between receiver and object & cm \\
        
        \textit{\(ToF\)} & Time of Flight & ns \\
        
        \textit{\(c\)} & Speed of light & (\(3.0 \times 10^8\)m/s) \\
        
        \textit{\(d_{TX\text{-}RX}\)} & Distance between the transmitter and receiver & cm \\
        \textit{\(p\)} & Point on ellipse indicating the target & - \\
        
        \textit{\(\alpha\)} & Phase difference between the two antennas on the same chip & radians \\
        
        \textit{\(\theta\)} & Angle of arrival & radians \\
        
        \textit{\(TP\)} & True positive, the peak that indicates a real object & - \\
        
        \textit{\(FP\)} & False positive, a peak retained as an object, but actually noise & - \\
        
        \textit{\(FN\)} & False negative, a peak corresponding to a real object that is filtered out & - \\
        
        \textit{\(P\)} & Precision calculated as TP / (TP + FP) & - \\
        
        \textit{\(R\)} & Recall calculated as TP / (TP + FN) & - \\
        \textit{F1-score} & Calculated as 2 x (P x R) / (P + R) & - \\
        \hline
        \hline
        \end{tabular}
    \end{center}
    \label{tab:mathematical-symbols}
\end{table}

\subsubsection{Signal processing}
\label{subsubsec:signal-processing}
To reliably extract the peak indicating an object, every raw CIR undergoes the following processing phases. Firstly, the CIR samples are initially recorded as I/Q components. To calculate the magnitude of the received samples, ~\eqref{eq:amplitude-calculation} is used on the Preamble CIR and ~\eqref{eq:phase-calculation} is used to calculate the phase of the received signals on both STS-based CIRs.

\begin{equation} \label{eq:amplitude-calculation}
    CIR_{magnitude} = \sqrt{I^2 + Q^2}
\end{equation}

\begin{equation}
    \phi = \arctan\left(\frac{Q}{I}\right)
    \label{eq:phase-calculation}
\end{equation}

The first four samples of the CIRs are excluded from further processing, as these samples precede the arrival of the first signal and represent the noise floor. However, these initial samples are used later for calculating the SNR-based score for filtering purposes.
Finally, to ensure a consistent threshold and comparability across different measurements, the preamble CIR is normalized using min-max scaling.

\subsubsection{Peak detection}
Peak detection is performed on the preamble CIR by identifying local maxima through comparison with neighbouring values. Subsequently, peaks are filtered based on their properties, as illustrated in the following step.

\subsection{Step 2: Filtering}
\label{subsec:step2}

\begin{figure}
    \centering
    \includegraphics[width=1\linewidth]{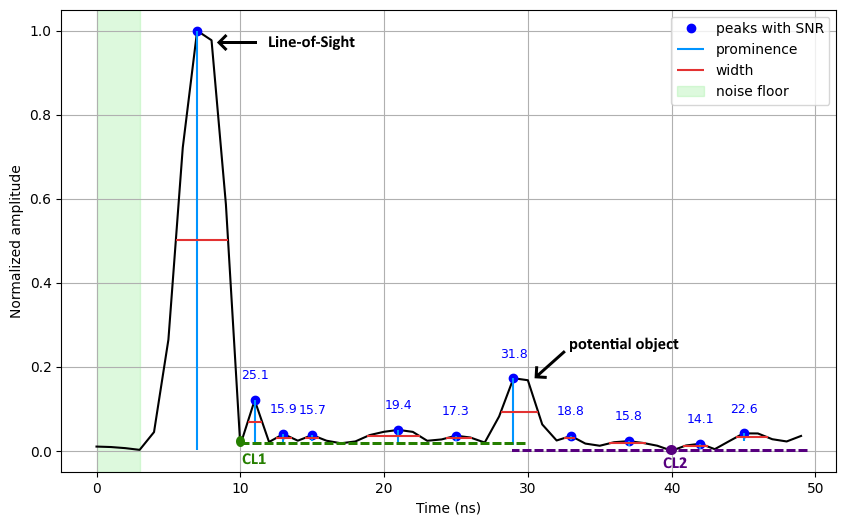}
    \caption{Definition of prominence, width, and the SNR-based score. Prominence is the vertical distance between a peak and its lowest contour line. In this example, the prominence of the potential object's peak is measured relative to contour line CL1 rather than CL2, since CL2 has a lower value. The width is calculated at half the prominence, using interpolation to determine the intersection points. The SNR-score evaluates the detected peak relative to the noise floor.}
    \label{fig:filtering}
\end{figure}

The filtering phase is applied to remove peaks that are likely to correspond to noise or unwanted multipath effects. This filtering process is based on four criteria:
\begin{itemize}
    \item \textit{prominence}: represents how much a peak stands out due to its amplitude and its location relative to other peaks. It is defined as the vertical distance between the peak and its lowest contour line (CL), as illustrated in Fig.\ref{fig:filtering}. The lowest CL is determined by searching to the left or right until a higher peak is found, or until the end of the CIR is reached. On each side, the minimum amplitude is taken, and the higher of the two values defines the peak's lowest CL. For the peak generated by a potential object, the prominence is measured between the peak and CL1. Since CL2 has a lower value, the contour line is CL1.
    \item \textit{width}: the minimum required width of a peak to be considered significant, calculated at half the prominence, with linear interpolation used to locate the intersection points (Fig. \ref{fig:filtering}). Narrow peaks are more likely to represent noise, whereas reflections from physical objects typically produce broader peaks.
    \item \textit{SNR-score\(_i\)}: the minimum required SNR-score for a detected peak to be considered a real object. The score is calculated as shown in (\ref{eq:snr}).
    \begin{equation}
        \text{\textit{SNR-score}}_{i} = 20\log_{10} \left( \frac{A_i}{A_\text{noise}} \right) + k \times delay_i
        \label{eq:snr}
    \end{equation}

    Where \( A_i \) is the amplitude of the detected peak (at index \textit{i}) and \( A_{noise} \) is calculated as the root mean square (RMS) of the noise floor (Fig. \ref{fig:filtering}). Additionally, to account for signal loss caused by the distance, a linear weighting is applied to the score based on a weighting factor \(k\) (0.20 in this case) and time delay \(delay_i\) between the peak index and the first path index. This adjustment helps detect objects farther away as they naturally have a lower signal amplitude.
    
    \item \textit{PDoA}: The antenna array's sensitivity and resolution degrade at wider angles, leading to increased multipath interference and reduced angular accuracy. The calculation of the PDoA, shown in ~\eqref{eq:phase_difference}, and adapted from \cite{pdoa-aoa}, no longer requires synchronization since both antennas are on the same chip.
    \begin{equation}
        \alpha = \left( (- \phi_A + \phi_B + \pi) \bmod 2\pi \right) - \pi
        \label{eq:phase_difference}
    \end{equation}
    
    Here \(\phi_A\) and \(\phi_B\) are the phases measured at antenna A and B.
        
    As such, peaks that have a corresponding PDoA value outside the range of -2.1325 to 2.1325 radians are discarded as estimates beyond this interval are considered unreliable. These bounds are derived using the inverse of \eqref{eq:angle_of_arrival}, which maps an AoA of -45$^{\circ}$ to 45$^{\circ}$ to the corresponding PDoA values.

\end{itemize}


\subsection{Step 3: Environment Mapping}
\label{subsec:noise-suppression}
The final stage of the proposed method combines the processed observations into a representation of the environment. This process involves three iterative steps: distance estimation, AoA determination and clustering. The distance estimation provides the range of each peak, whereas the AoA is used to position the detected peaks relative to the robot. Finally, peaks originating from the same location are grouped through clustering.
\subsubsection{Distance Estimation}
For obstacle detection, the peak with the highest amplitude (excluding the first path) that meets all the conditions is selected, as it is most likely to correspond to a real object. This is illustrated in Fig. \ref{fig:CIRs}. Each peak beyond the first path arrives later because the reflected signal travels a longer time than the direct path. Therefore, the ToF of the reflected signal is used to measure the total distance between transmitter (TX), object, and receiver (RX).

\begin{figure}
    \centering
    \includegraphics[width=1\linewidth]{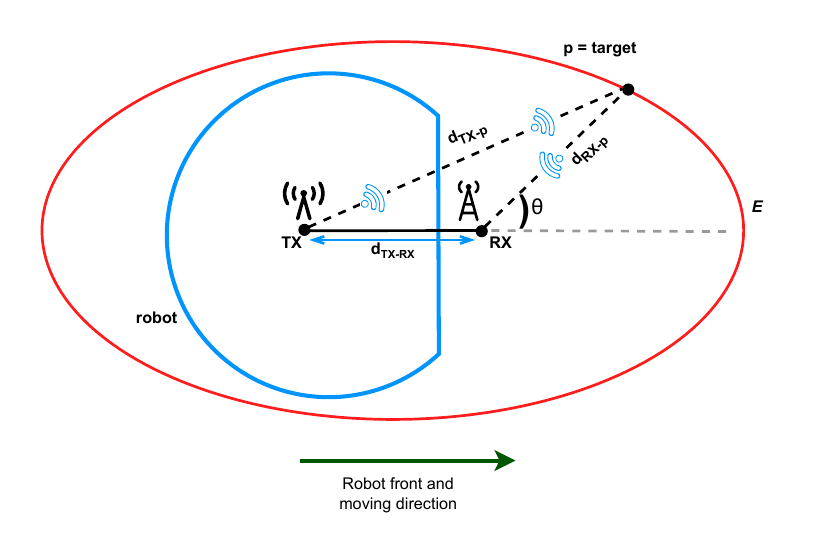}
    \caption{Bistatic setup: The ellipse \(E\) is defined by the total path length (\(d_{TX-p} + d_{RX-p}\)) and the fixed positions of the TX and RX antennas, used as the foci of the ellipse. \(d_{TX\text{-}RX}\) is the distance between the TX antenna and RX antenna, \(d_{TX-p}\) is the distance between TX antenna and target \(p\), \(d_{RX-p}\) is the distance between target and RX antenna, and \(\theta\) is the AoA .}
    \label{fig:ellips_distances}
\end{figure}

Fig. \ref{fig:ellips_distances} illustrates an ellipse \(E\) whose foci are the TX antenna and RX antenna. Each point \(p\) on the ellipse represents a possible target and is defined such that the sum of the distances \(d_{TX-p}\) and \(d_{RX-p}\) equals the total path length travelled by the signal between TX, target, and RX, as in \eqref{eq:ellipse-distance}. The total path length equals the ToF multiplied by the speed of light \(c\). To resolve the exact location and distance relative to the robot, the angle of arrival must additionally be determined, as described in \eqref{eq:angle_of_arrival}.

\begin{equation}
    E = {p\in 	\mathbb{R}^2: d_{TX-p} + d_{RX-p} =  ToF \times c}
    \label{eq:ellipse-distance}
\end{equation}

\subsubsection{Angle of Arrival}
\label{subsubsec:AoA}

To estimate the AoA, the PDoA between the two antennas on the same chip is used. This is calculated as shown in ~\eqref{eq:phase_difference}, followed by ~\eqref{eq:angle_of_arrival}, which shows the calculation of the AoA \cite{pdoa-aoa}.

\begin{equation}
    \theta = \frac{1}{0.95}\arcsin\left(\frac{\alpha}{\pi}\right)
    \label{eq:angle_of_arrival}
\end{equation}
Tests show that the estimated AoA generally follows the ground truth when the object is in front of the robot. Some deviations appear, especially when approaching the object from the side, but these estimates remain useful for filtering.

The distance to the RX antenna can be calculated using the cosine law on the formed triangle (Fig. \ref{fig:ellips_distances}) and the measured total path length.

\begin{equation}
    d_{TX-p}^2 = d_{RX-p}^2 + d_{TX\text{-}RX}^2 - 2d_{RX-p}d_{TX\text{-}RX}\cos\theta
\end{equation}

This equation contains two unknown distances, \(d_{TX-p}\) and \(d_{RX-p}\), and can be transformed to \eqref{eq:distance-rx}.

\begin{equation}
    d_{RX-p} = \frac{d^2 - d_{TX\text{-}RX}^2}{2(d + d_{TX\text{-}RX}\sin\theta)} 
    \label{eq:distance-rx}
\end{equation}

Here, \(d_{TX\text{-}RX}\) denotes the distance between the TX antenna and RX antenna, \(d\) is the total path length (\(d_{TX-p}\) + \(d_{RX-p}\)), and \(\theta\) is the calculated AoA. This equation calculates the distance between the object and the RX antenna.

\subsubsection{Clustering}
\label{subsubsec:clustering}
By applying the filtering methods from Section \ref{subsec:step2} and localising the resulting points relative to the robot using the distance and orientation calculations, a point cloud of potential obstacles is generated. In this final step, a clustering algorithm then removes outliers and groups these points into likely obstacle regions. To this end, the \textit{Density-Based Spatial Clustering of Applications with Noise (DBSCAN)} is used.
DBSCAN is well suited for this work because detected components from real objects tend to appear repeatedly in the same location, forming dense groups of points. By identifying these dense regions, DBSCAN can cluster detected components that originate from the same physical object while simultaneously discarding isolated points as noise.

\begin{figure}
    \centering
    \includegraphics[width=1\linewidth]{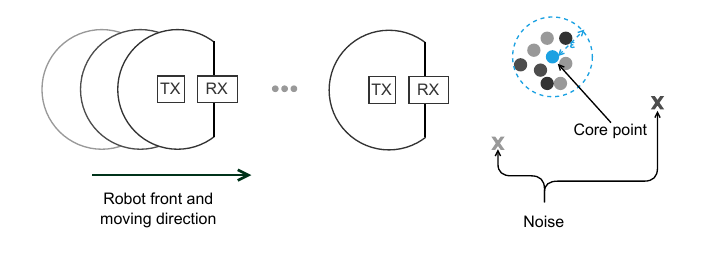}
    \caption{Illustration of the clustering process: Peaks detected across multiple robot positions that form dense regions are grouped into clusters and noise is removed.}
    \label{fig:clustering-overview}
\end{figure}

\rev{The key DBSCAN parameters used are (see Fig. \ref{fig:clustering-overview}):}
\begin{itemize}
    \item \rev{\textit{epsilon (eps)}: the maximum distance between two samples for them to be considered neighbours.}
    \item \rev{\textit{min\_samples}: the number of samples required in a neighbourhood for a point to be considered a core point, including the point itself.}
\end{itemize}
\rev{Clusters that do not meet these thresholds are discarded as noise. In this paper, \textit{eps} was set to 20 cm and \textit{min\_samples} to 20 detected peaks. Using smaller values would reduce the number of points labelled as noise but increases the risk of forming incorrect clusters, which can degrade mapping accuracy.} Clustering is performed based on peaks mapped to the Cartesian coordinate system and depends on the number of detected peaks across multiple measurements (these could correspond to different robot positions, although the robot may also remain in the same position (see Fig. \ref{fig:clustering-overview})).

Since this work aims to establish a foundation for anchor-free UWB-based obstacle mapping that can be integrated into broader SLAM systems, autonomous navigation or real-time obstacle avoidance capabilities are not yet incorporated.

\section{Data Collection}
\label{sec:data-collection}
\begin{figure}
    \centering
        \subfloat{\includegraphics[width=0.45\linewidth]{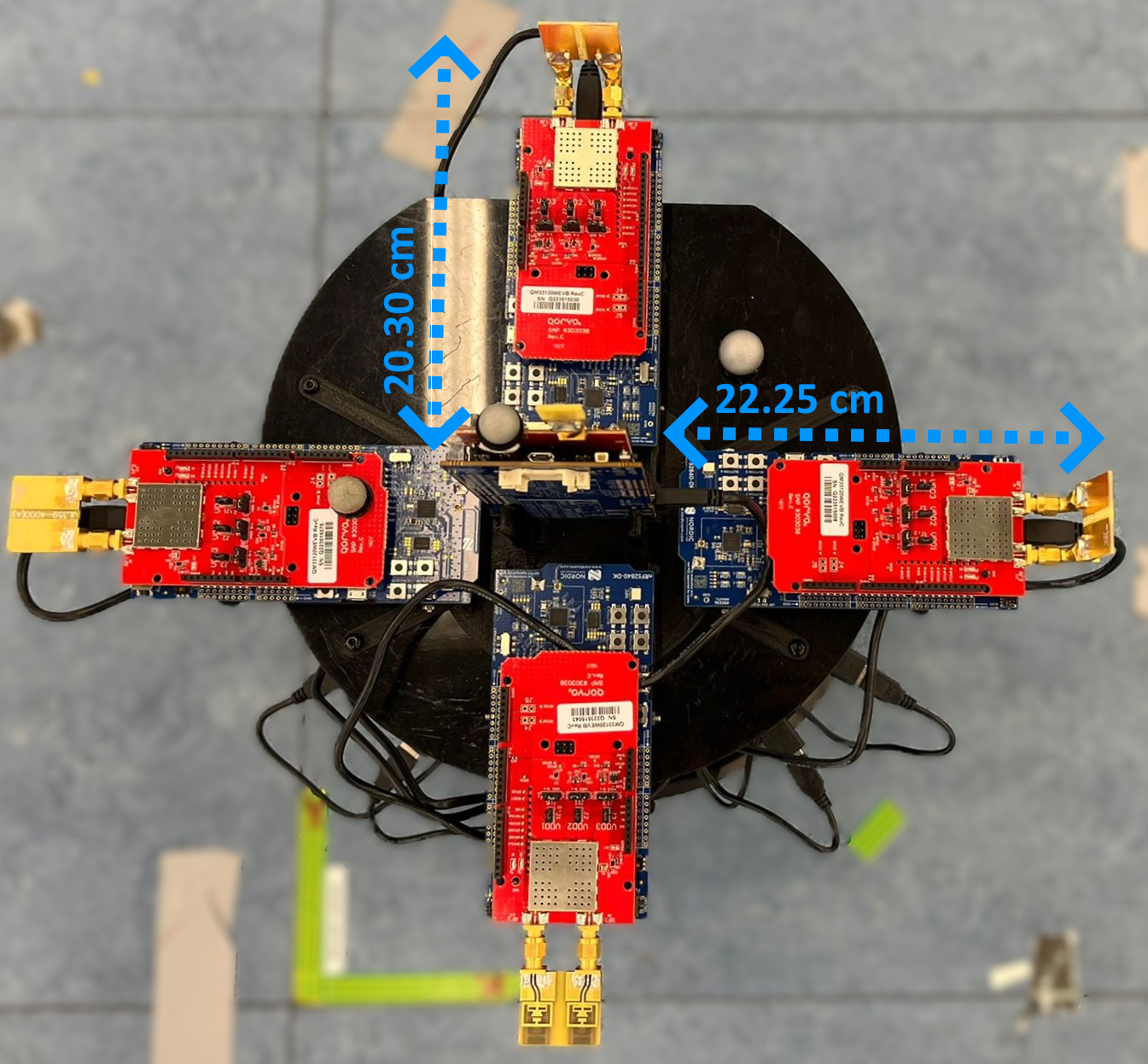}
        \label{fig:robot-antennas}}
    \hfil
        \subfloat{\includegraphics[width=0.45\linewidth]{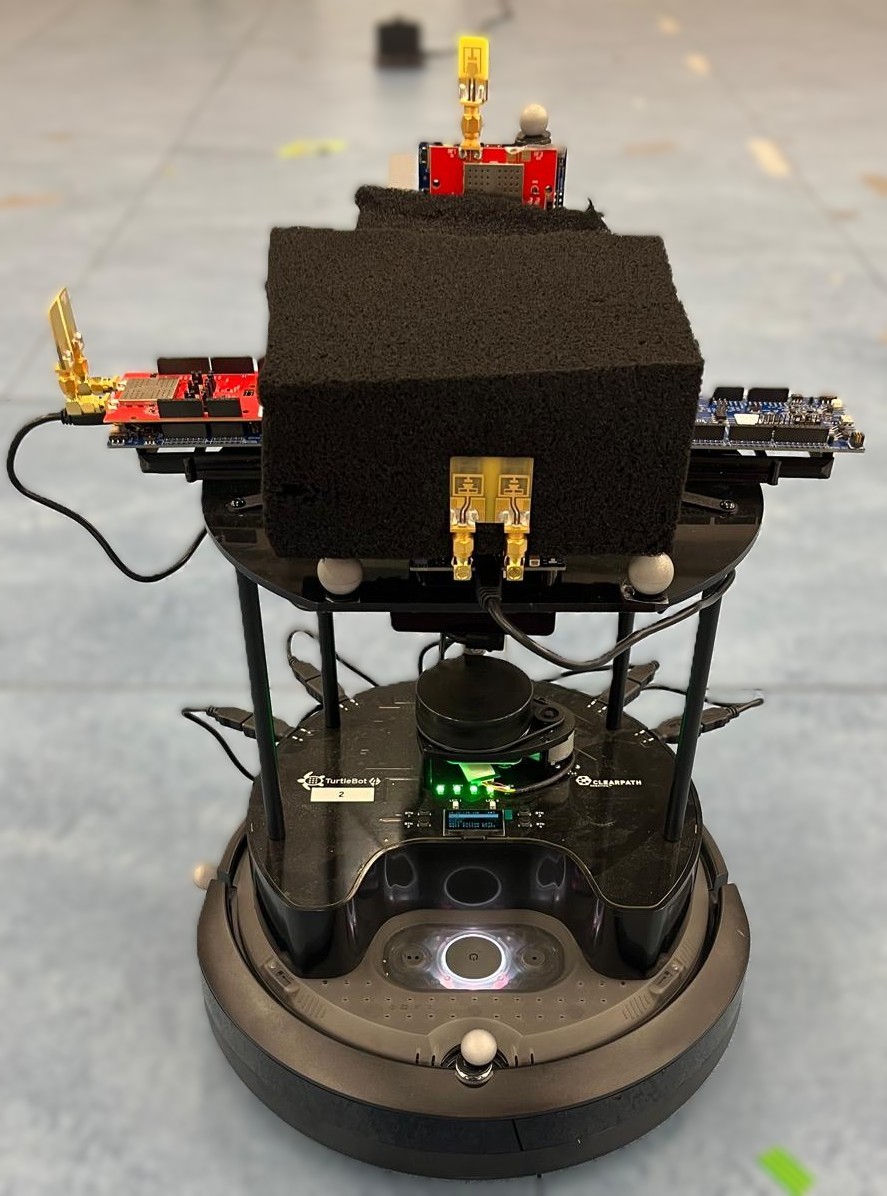}
        \label{fig:robot-absorber}}
    \caption{a) Overview of the antenna placement on the TurtleBot 4 including the distance between the transmitter and receivers.  b) Front view of the TurtleBot 4, equipped with the antennas and absorber.}
    \label{fig:robot-overview}
\end{figure}

\rev{The following data collection experiments were conducted to evaluate the effectiveness of the proposed approach and assess its performance under realistic conditions. The data collection experiments were conducted in the controlled Industrial Internet of Things (IIoT) lab at Ghent University \cite{lab-ugent}.} All experiments were performed using a TurtleBot 4 robot, equipped with one omnidirectional UWB transmitter on the top and multiple dual-antenna, directional UWB receivers on every side depending on the experiment (Fig. \ref{fig:robot-antennas}). \rev{Because all antennas are connected to the same onboard computer of the robot, each receiver is synchronised with millisecond accuracy.} To reduce the impact caused by the first path signal as the transmitter is placed close to the receiver, an absorber is used (Fig. \ref{fig:robot-absorber}). The specifications of the radar system are listed in Table \ref{tab:uwb-specifications}. During data capture, the system alternated between channel 5 and channel 9 for each measurement to assess signal behaviour across different frequency bands. The IIoT lab is equipped with a Motion Capture (MOCAP) system, providing millimetre-level accuracy ground truth data as the robot moves.

\begin{table}
    \caption{\rev{UWB Radar Configurations - IEEE 802.15.4 compliant}}
        \begin{center}
        \renewcommand{\arraystretch}{1.2}
        \begin{tabular}{ll}
        \hline
        \textbf{Parameter} & \textbf{Value} \\
        \hline
        \hline
        UWB sensor & QM33120WDK1 \cite{datasheet-sensors} \\
        Preamble length & 512 \\
        Frequency & 6.5 GHz and 8.0 GHz \\
        Channel & 5 and 9 \\
        Bandwidth & 500 MHz \\
        Transmitter specification & JL159 Omnidirectional single antenna\\
        Receiver specification & JL359 Directional dual antenna\\
        Update rate & 96 Hz \\
        Number of I/Q samples & 50 I/Q pairs\\
        \rev{Fast-time sampling interval} & \rev{1 ns} \\
        \hline
        \hline
        \end{tabular}
    \end{center}
    \label{tab:uwb-specifications}
\end{table}

\begin{figure}
    \centering
        \subfloat{\includegraphics[width=0.3\linewidth]{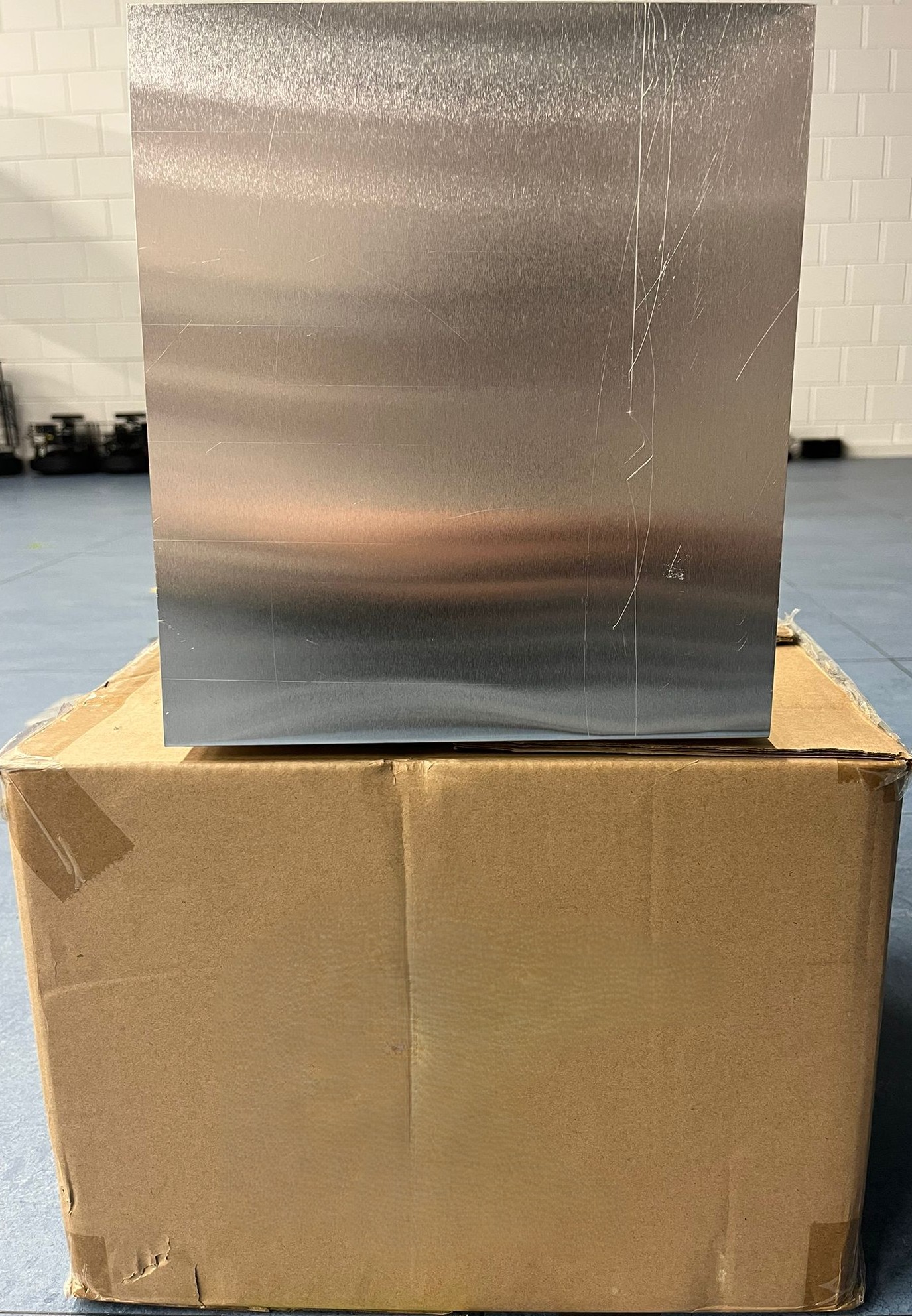}
        \label{fig:metal-plate}}
    \hfil
        \subfloat{\includegraphics[width=0.3\linewidth]{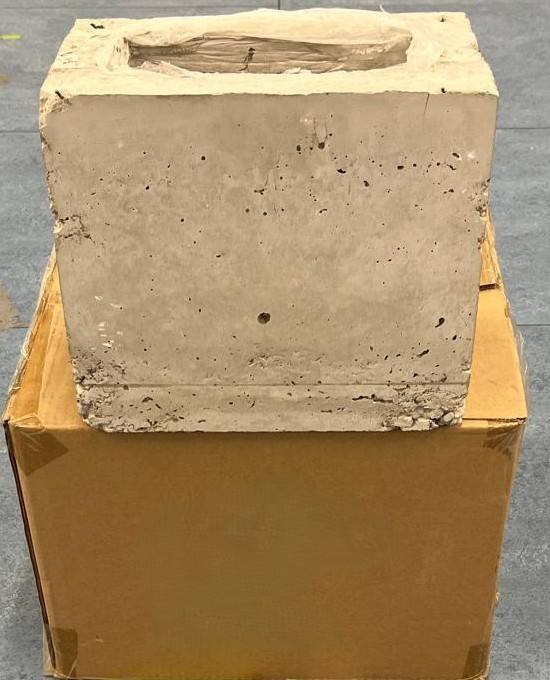}
        \label{fig:concrete-box}}
    \hfil
        \subfloat{\includegraphics[width=0.3\linewidth]{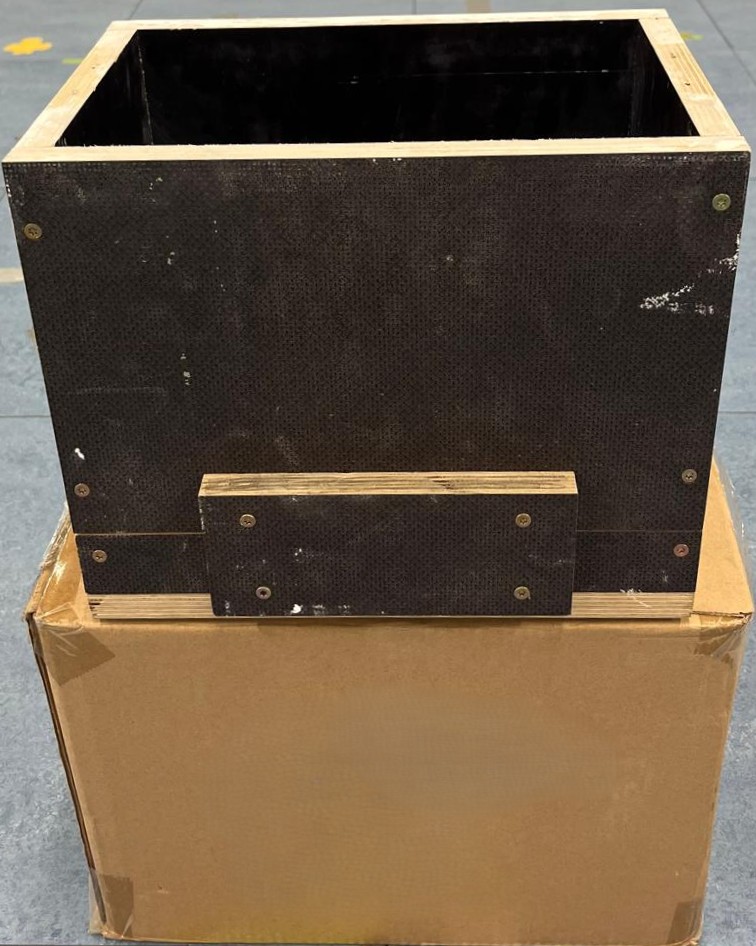}
        \label{fig:plywood-box}}
    \caption{a) metal plate measuring 25 cm x 28 cm b) concrete box measuring 28 cm x 24 cm c) plywood box measuring 31.5 cm x 26 cm}
    \label{fig:materials}
\end{figure}

\begin{figure}
    \centering
        \subfloat[\rev{First data collection}]{\includegraphics[width=0.90\linewidth]{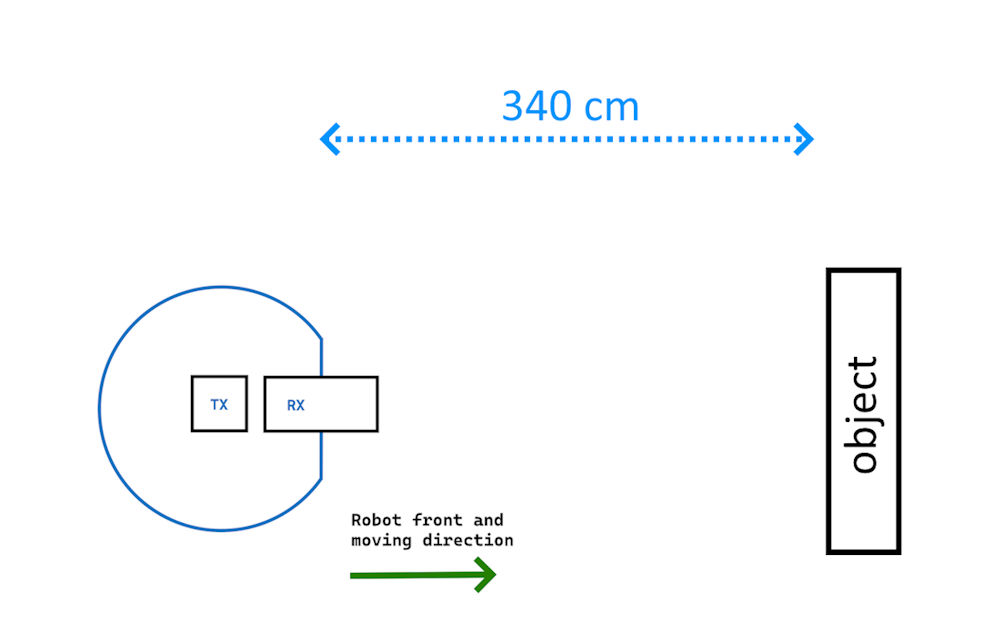}
        \label{fig:first-data-collection}}
    \hfil
        \subfloat[\rev{Second data collection}]{\includegraphics[width=0.90\linewidth]{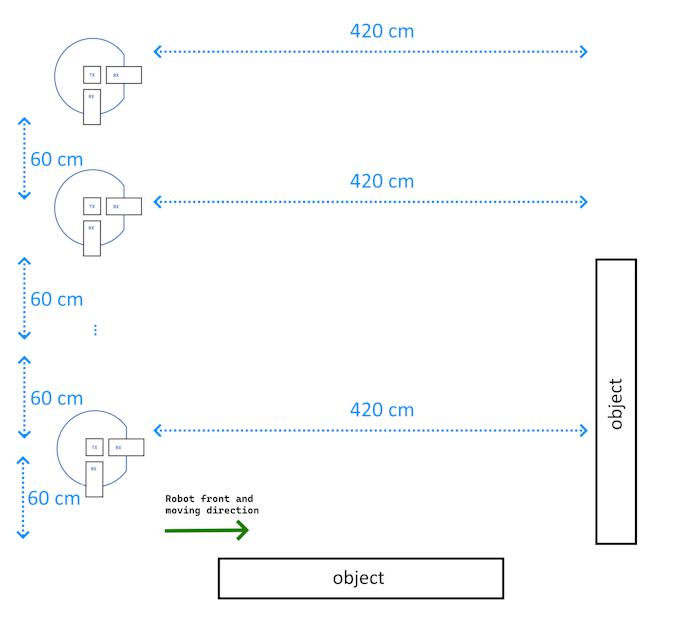}
        \label{fig:second-data-collection}}
    \caption{\rev{a) The robot moves straight forward to the object. The same movement is done three times, once for every material (metal, concrete, plywood). b) Two objects are placed, in the front and parallel to the robot while data is captured using two receivers.}}
    \label{fig:data-collections}
\end{figure}

\subsection{First data collection: different materials}
\label{subsec:first-data-collection}
The goal of this data collection was to analyse how UWB radar signals behave depending on the type of material. Three different objects were used in the experiments: a metal plate measuring 25 cm x 28 cm (Fig. \ref{fig:metal-plate}), a concrete box measuring 28 cm x 24 cm (Fig. \ref{fig:concrete-box}), and a plywood box measuring 31.5 cm x 26 cm (Fig. \ref{fig:plywood-box}). These materials were selected for their varying electromagnetic properties: the metal plate is highly reflective, the concrete box partially reflects and absorbs signals, and the plywood box primarily absorbs with minimal reflection. As shown in Fig. \ref{fig:first-data-collection}, the robot moved in a straight line toward a single object, starting from a distance of 340 cm. For this experiment, only the top transmitter and front receiver were active. Due to the use of different materials, this setup is a good validation of the accuracy of recognizing different material types. 

\subsection{Second data collection: multiple objects}
\label{subsec:second-data-collection}
The second data collection aimed to explore the ability to detect objects not placed directly in front of the receiver. In this setup, two white plates made of highly reflective material with dimensions of 116 cm by 55 cm were placed as target objects, one directly in front of the robot and the other parallel to its path to test if the front antenna could detect signals from the side or vice versa (Fig. \ref{fig:second-data-collection}). The robot started from a distance of 420 cm from the front plate and 60 cm from the parallel plate. The distance to the parallel object increased in 60 cm steps up to 300 cm, while the robot was repositioned at the same start point. Since both objects are highly reflective, this setup creates many multi-path components and phantom peaks and is hence a good validation of the efficiency of the filtering algorithm. 

\section{Results}
\label{sec:results}
This section presents and analyses the results from the proposed processing pipeline. It begins with an overview of the filtering parameters, followed by an analysis of target identification, filtering performance and clustering behaviour. This section concludes with a visual demonstration to validate all subcomponents.

The performance of the proposed approach is evaluated using two types of metrics: (i) detection probability and (ii) distance accuracy (expressed as distance error to indicate how well obstacles are positioned relative to the robot). 

Detection probability is expressed as Precision \textit{P}, Recall 
\textit{R}, and F1-score, as defined in ~\eqref{eq:precision}-\eqref{eq:f1_score}:

\begin{equation}
    P = \frac{TP}{TP + FP} 
    \label{eq:precision}
\end{equation}

\begin{equation}
    R = \frac{TP}{TP + FN} 
    \label{eq:recall}
\end{equation}

\begin{equation}
    \text{\textit{F1-score}} = 2 \frac{\textit{PR}}{P + R}
    \label{eq:f1_score}
\end{equation}

Where \textit{TP} (true positives) are peaks correctly retained that correspond to actual objects, \textit{FP} (false positives) wrongly kept noise peaks that do not correspond to real objects (e.g. walls or noise), and \textit{FN} (false negatives) are peaks incorrectly removed that do correspond to actual objects. For detection probability, a margin of 20 cm is used when comparing estimated distances to ground truth: an estimated peak is considered a true positive if it lies within 20 cm of the ground truth object.

Distance accuracy was evaluated in two different ways. First, after target identification and filtering, the detected peaks were assumed to lie directly in front of the antenna, since no AoA information was available. In this case, distance error was computed as the mean absolute error (MAE) between the ground truth and the estimated distance. The results are discussed in Section \ref{subsubsec:impact-filtering}.

Second, when AoA information was included, the retained peaks could be mapped relative to the robot's position, after which clustering was applied. The distance error was then calculated as the shortest distance between the ground truth coordinates and the estimated object coordinates in the x/y plane. Finally, the mean of these values was taken. The results are discussed in Section \ref{subsubsec:impact-mapping}.

\subsection{Filtering parameters}
\label{subsec:filtering-params}
\rev{The three key parameters (peak \textit{width}, \textit{prominence}, and \textit{SNR-score}) were optimised using the dataset obtained from the first data collection (Fig. \ref{fig:first-data-collection}). A grid search over different parameter values was performed, and each configuration was evaluated using the \textit{F1-score} and the root mean square error (RMSE) between the estimated and ground-truth distances. The parameters achieving the highest \textit{F1-score} and lowest RMSE were selected. The optimised values are summarised in Table \ref{tab:parameter-results}, both for specific materials and overall optimal values across different materials. These parameters do not need to be tuned to these exact values. Our analysis shows that a zone exists in the parameter space in which the detection quality remains consistently high, even when the parameters are varied slightly. This stable high-performance zone is noticeably larger for highly reflective materials, whereas less reflective materials such as concrete require parameters closer to their optimal values due to increased sensitivity.} 

\begin{figure}
    \centering
        \subfloat[\rev{Prominence decreases with distance for all materials}]{\includegraphics[width=0.90\linewidth]{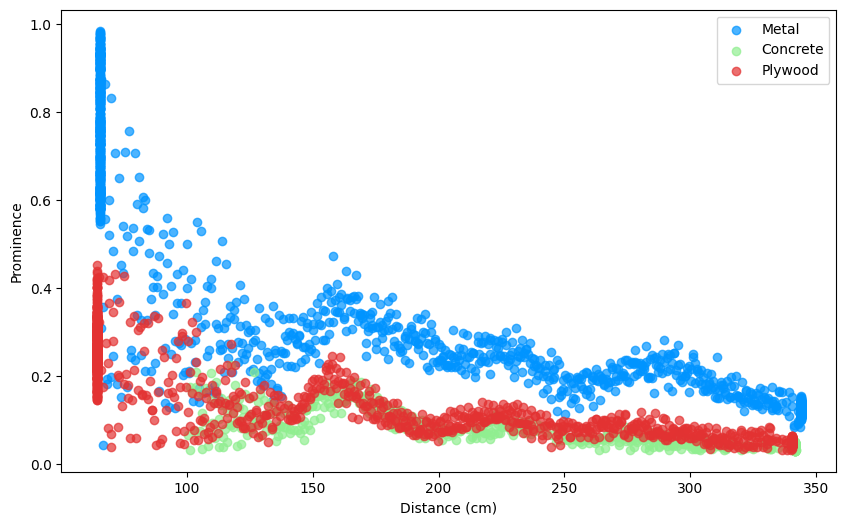}
        \label{fig:prominence-vs-distance}}
    \hfil
        \subfloat[\rev{Width increases with distance for all materials}]{\includegraphics[width=0.90\linewidth]{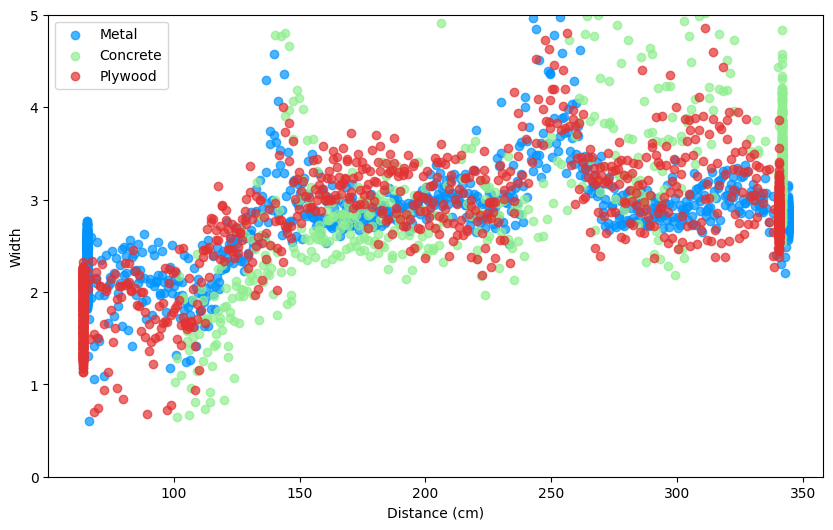}
        \label{fig:width-vs-distance}}
    \caption{\rev{Correlation between peak properties and distance.}}
    \label{fig:params-vs-distance}
\end{figure}

To further investigate their behaviour, the relationship between distance and both peak prominence and width was analysed for the three different materials (metal, concrete, plywood). Figures \ref{fig:prominence-vs-distance} and \ref{fig:width-vs-distance} illustrate the correlations in peak prominence and width as a function of target distance.

As seen in Fig. \ref{fig:prominence-vs-distance}, peak prominence decreases as the distance increases. At short ranges, the reflected signals retain more energy and produce stronger peaks. With increasing distance, signal attenuation and energy loss cause the peaks to stand out less compared to other peaks or the noise floor. Metal consistently produces the highest peak prominence, which is expected given its high reflectivity. Concrete and plywood, by comparison, show lower prominence because they absorb more energy. At distances lower than 100 cm, concrete cannot be detected with the current setup as the peak prominence and width are too low. As distance grows, both concrete and plywood follow a similar downward trend. This limitation is not exclusive to this material. It can affect any material type depending on its reflectivity. Highly reflective surfaces like metal return more energy producing more prominent peaks even when close to the robot. In contrast, materials such as concrete and plywood absorb more energy, resulting in weaker reflections that are more easily overshadowed by the strong first path signal. When the object is within approximately 60 cm, the reflected peak becomes difficult to distinguish due to the strong first path signal and oversaturation of the receiver \cite{comparative-lidar-radar}. Although an absorber was used between the antennas to mitigate this issue, the results show that the problem persists indicating that antenna design also plays a crucial role. This could be resolved by the design of a dedicated UWB antenna array with better isolation between the individual antenna components \cite{uwb-antenna}.

Fig. \ref{fig:width-vs-distance} shows that the peak width increases with distance for all three materials. Close to the robot, the peaks are relatively narrow (around 1-2 ns), but at greater distances they broaden to approximately 3-4 ns. The time delay makes the signal arrive over a longer period, which appears as a broader peak in the CIR.

\rev{The remainder of this section uses the overall filtering parameters listed in Table \ref{tab:parameter-results}. These parameters represent the best configuration for mixed-material environments containing metal, concrete, and plywood. If the material in the environment is known beforehand (i.e., only one material type is present), parameters specific to that material type can be used to further improve detection performance.}

\begin{table}
    \caption{Optimal peak width, peak prominence and SNR-score values for different materials and multiple radio channels.}
    \begin{center}
    \renewcommand{\arraystretch}{1.2}
    \begin{tabular}{c|lccc}
    \hline
    \textbf{Channel} & \textbf{Material} & \textbf{Width} & \textbf{Prominence} & \textbf{SNR-score} \\
    \hline \hline
    
    \multirow{4}{*}{\makecell{5}} & Metal    & 1    & 0.04 & 25 \\
    
                                          & Concrete & 1    & 0.05 & 20 \\
                                          
                                          & Plywood  & 1    & 0.02 & 15 \\
                                          
                                          & Overall  & 1    & 0.05 & 20 \\
    \hline
    
    \multirow{4}{*}{\makecell{9}} & Metal    & 1    & 0.05 & 20 \\
    
                                          & Concrete & 2 & 0.03 & 10 \\
                                          
                                          & Plywood  & 0.10 & 0.01 & 10 \\
                                          
                                          & Overall  & 0.20 & 0.03 & 10 \\
    \hline
    \hline
    \end{tabular}
    \end{center}
    \label{tab:parameter-results}
\end{table}

\subsection{Target identification, filtering, and mapping accuracy}
\subsubsection{Impact of different materials after filtering}
\label{subsubsec:impact-filtering}

\rev{The performance of the algorithm after target identification and filtering (on width, prominence, SNR-score, and PDoA) differed across materials and channels. The used parameters are illustrated in Table \ref{tab:parameter-results}. The distance accuracy and detection probability results are shown in Table. \ref{tab:distance-detection-step2}. In this step, AoA was not considered, and all detections were assumed in front of the antenna.}

\rev{\textbf{Metal}: The metallic plate was detected with consistently high accuracy on both channels. The strong reflectivity produced clear, distinct peaks, resulting in high precision and recall. Channel 9 further improved detection and distance accuracy.}

\textbf{Concrete}: Detection was more challenging for the concrete box. On channel 5, the object was only detected starting at 250 cm, with higher variability (MAE 19.92 cm, SD 36.96 cm) and reduced precision ($81.25\%$). Recall remained high ($93.77\%$), but the large number of outliers increased the error. At close range (<100 cm), the signal was overshadowed by the first path, and the object was not detected on both channels 5 and 9. In contrast, channel 9 considerably improved performance: detection began at 340 cm, MAE dropped to 10.50 cm, and precision increased to $90.71\%$. Fig. \ref{fig:prominence-vs-distance} and Fig. \ref{fig:width-vs-distance} illustrate that concrete is overshadowed at close distances because the width and prominence are relatively low.

\textbf{Plywood}: Results on channel 5 were the weakest, with low precision ($79.30\%$) and moderate recall ($68.04\%$). This is primarily due to the low amplitude of the peak, which is often close to the noise floor and difficult to differentiate from background noise. More reliable detection occurred when the robot was within approximately 120 cm. Channel 9 improved performance noticeably: MAE decreased to 7.99 cm (from 30.73 cm), and both precision ($98.14\%$) and recall ($99.23\%$) were near perfect, with consistent detection from 340 cm onward.

Overall, channel 9 consistently outperformed channel 5 for non-metallic materials, resulting in lower MAE, fewer outliers, and higher precision/recall. In addition, further accuracy gains can be obtained during step 3 of the processing pipeline (see next sections).

\begin{table*}
    \caption{\rev{After step 2: Distance accuracy (in cm) and detection probability (\%) after filtering. MAE denotes the mean absolute error, SD the standard deviation, and P90/P95 the 90th and 95th percentiles of the absolute distance error, respectively.}}
    \begin{center}
    \renewcommand{\arraystretch}{1.2}
    \begin{tabular}{c|lcccccccc}
    \hline
    \hline
    \multicolumn{2}{c}{} 
        & \multicolumn{4}{c|}{\textbf{Distance Accuracy (in cm)}} 
        & \multicolumn{3}{c}{\textbf{Detection Probability (\%)}} \\
        \cline{3-6} \cline{7-9}
    \textbf{Channel} & \textbf{Material} & \textbf{MAE} & \textbf{P90} & \textbf{P95} & \textbf{SD} & \textbf{Precision} & \textbf{Recall} & \textbf{F1\textit{-}score}\\
    
    \multirow{3}{*}{\makecell{5}} 
         & Metal    & 6.49  & 12.34 & 12.34 & 14.25 & 98.93 & 100.00 & 99.46 \\
         & Concrete & 19.92 & 44.24 & 116.62 & 36.96 & 81.25 & 93.77 & 87.06 \\
         & Plywood  & 30.74 & 48.02 & 231.51 & 63.08 & 79.30 & 68.04 & 73.24 \\
    \hline
    
    \multirow{3}{*}{\makecell{9}}
                 & Metal    & 5.80 & 14.33 & 14.34 & 4.70 & 99.67 & 100.00 & 99.83 \\
                 & Concrete & 10.50 & 18.49 & 33.32 & 11.37 & 90.71 & 88.40 & 89.54 \\
                 & Plywood  & 7.99 & 18.81 & 18.82 & 7.40 & 98.14 & 99.23 & 98.68 \\
    \hline
    \hline
    \end{tabular}
    \end{center}
    \label{tab:distance-detection-step2}
\end{table*}

\begin{table*}
    \caption{\rev{After step 3: Distance accuracy (in cm) and detection probability (\%) after filtering, angle of arrival, and mapping. MAE denotes the mean absolute error, SD the standard deviation, and P90/P95 the 90th and 95th percentiles of the absolute distance error, respectively.}}
    \begin{center}
    \renewcommand{\arraystretch}{1.2}
    \begin{tabular}{c|lcccccccc}
    \hline
    \hline
    \multicolumn{2}{c}{} 
        & \multicolumn{4}{c|}{\textbf{Distance Accuracy (in cm)}} 
        & \multicolumn{3}{c}{\textbf{Detection Probability (\%)}} \\
        \cline{3-6} \cline{7-9}
    \textbf{Channel} & \textbf{Material} & \textbf{MAE} & \textbf{P90} & \textbf{P95} & \textbf{SD} & \textbf{Precision} & \textbf{Recall} & \textbf{F1\textit{-}score}\\
    
    \multirow{3}{*}{\makecell{5}} 
                    & Metal    & 8.21  & 16.63 & 23.76 & 7.49 & 93.20 & 99.93 & 96.44\\
                    & Concrete & 9.01  & 18.43 & 24.18 & 6.82 & 93.73 & 91.55 & 92.63\\
                    & Plywood  & 11.24 & 19.71 & 29.65 & 8.63 & 93.31 & 79.26 & 85.71\\
    \hline
    
    \multirow{3}{*}{\makecell{9}}
                     & Metal    & 6.85 & 14.73 & 17.40 & 5.87 & 96.93 & 99.11 & 98.01\\
                     & Concrete & 16.89 & 39.19 & 46.47 & 13.90 & 73.42 & 83.38 & 78.08\\
                     & Plywood  & 8.47 & 18.87 & 18.90 & 6.96 & 98.55 & 99.10 & 98.82\\
    \hline
    \hline
    \end{tabular}
    \end{center}
    \label{tab:distance-detection-step3}
\end{table*}

\subsubsection{\rev{Impact of combined filtering and mapping}}
\label{subsubsec:impact-mapping}

\begin{table}
    \caption{\rev{Key parameters used in step 3 (filtering + mapping).}}
    \begin{center}
    \renewcommand{\arraystretch}{1.2}
    \begin{tabular}{lP{1.9cm}P{3.5cm}}
    \hline
    \textbf{Parameter} & \textbf{Threshold} & \textbf{Comments} \\
    \hline \hline
    \textit{AoA\_range} & [\ang{-45},\ang{45}] & Valid AoA interval.\\
    
    \textit{eps} & 20 cm & Max distance between two samples to be part of the same neighbourhood (DBSCAN).\\
    
    \textit{min\_samples} & 20 & Number of samples in a neighbourhood for a point to be considered as a core point (DBSCAN).\\
    
    \textit{min\_peaks} & 50 & Minimum number of new peaks before clustering. \\
    
    \textit{range} & 20 cm & True positive distance range. \\
    
    \rev{\textit{bias}} & \rev{ch5 - 15 cm} & \rev{Distance calibration offset.} \\
    \rev{\textit{bias}} & \rev{ch9 - 13 cm} & \\
    \rev{\textit{bias\_AoA}} & \rev{ch5 - 0.0522 rad} & \rev{Angular calibration offset.} \\
    \rev{\textit{bias\_AoA}} & \rev{ch9 - 0.0209 rad} &\\
    \hline
    \hline
    \end{tabular}
    \end{center}
    \label{tab:used-param}
\end{table}

\rev{After AoA estimation, the detected peaks are no longer assumed to be located directly in front of the robot. Instead, their positions are calculated relative to the robot and mapped onto the x/y plane. The UWB chips used in the proposed approach have an angular measurement accuracy of approximately \ang{5}. As a result, differences can occur between the true and estimated angle of arrival, depending on the channel and material used, as explained later in this section. At larger distances, even small AoA estimation errors result in significant position errors.}

\rev{The filtering parameters from Table \ref{tab:parameter-results} and mapping parameters from Table \ref{tab:used-param} are used.} \rev{Table \ref{tab:distance-detection-step3} shows that the detection performance remains largely unchanged after this step. Precision and recall slightly increase for plywood, while a decrease is observed for concrete on channel 9, with an improvement on channel 5. This confirms that AoA accuracy is influenced by both channel selection and material properties.}
\rev{A clear reduction in standard deviation (SD) is observed across nearly all channel-material combinations, indicating a lower spread in the error distribution. This confirms that the clustering step effectively removes spatially inconsistent detections that do not belong to a valid cluster. In general, the remaining distance errors stay below 30 cm, except for concrete on channel 9, where larger AoA deviations lead to increased errors.}

\subsubsection{\rev{Complete pipeline visualisation}}

\begin{figure*}
    \centering
        \subfloat[]{\includegraphics[width=0.32\linewidth]{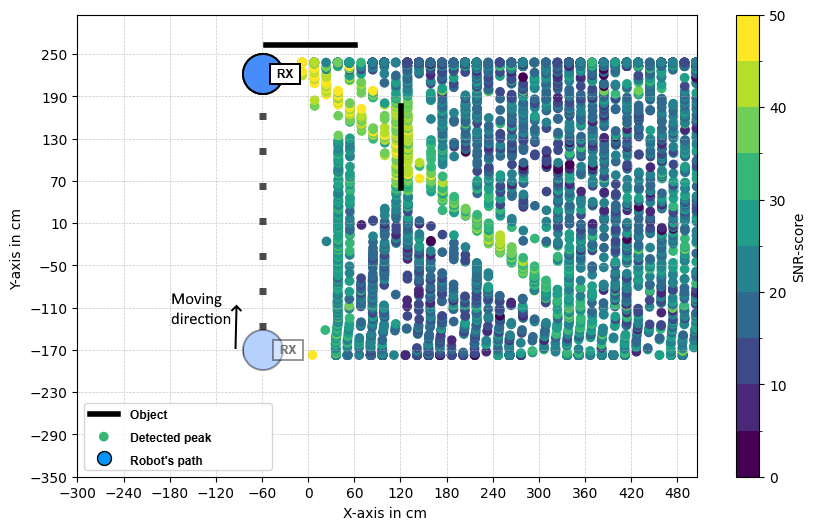}
        \label{fig:step-0}}
    \hfil
        \subfloat[]{\includegraphics[width=0.32\linewidth]{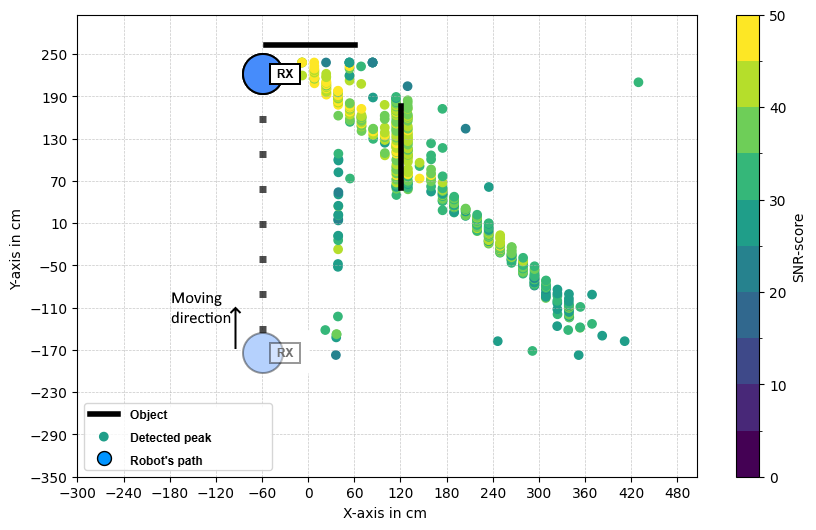}
        \label{fig:step1}}
    \hfil
        \subfloat[]{\includegraphics[width=0.32\linewidth]{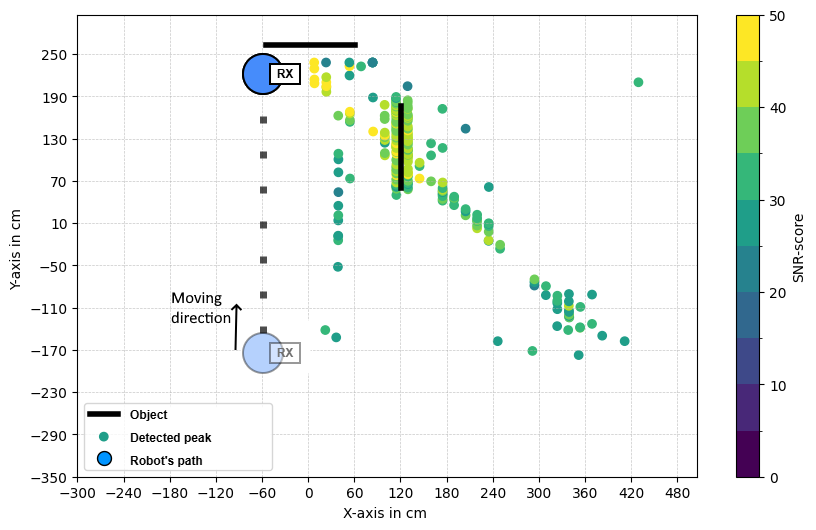}
        \label{fig:step2}}
    \hfil
        \subfloat[]{\includegraphics[width=0.45\linewidth]{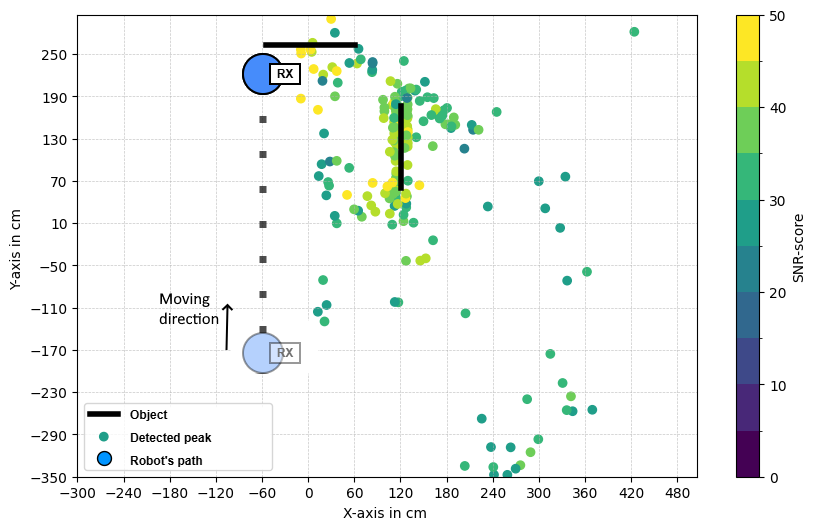}
        \label{fig:step3}}
    \hfil
        \subfloat[]{\includegraphics[width=0.45\linewidth]{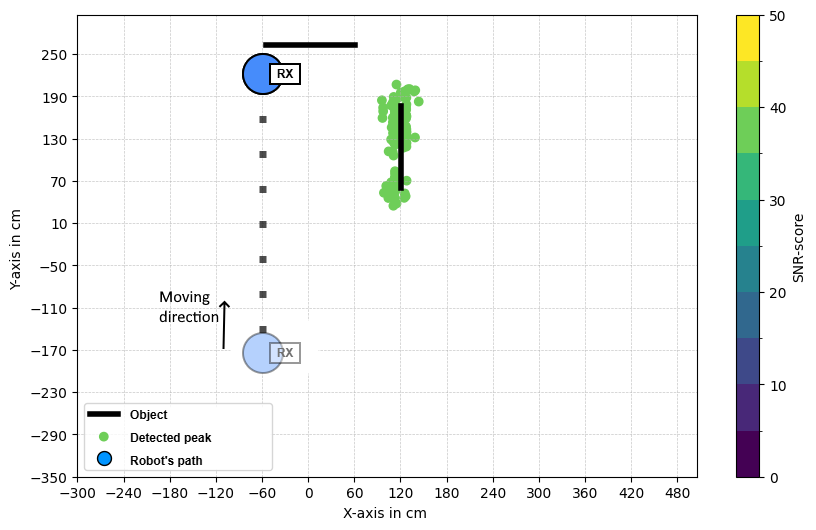}
        \label{fig:step4}}
    \caption{Overview of the impact of each of the multi-stage processing pipeline steps for object detection. For visual clarity, only the point cloud from the right-side antenna is shown. a) Peak detection: 5601 initial peaks b) Filtering on \textit{width}, \textit{prominence} and \textit{SNR-score}: 444 remaining peaks c) Additional filtering on PDoA: 225 remaining peaks d) Calculating object position relative to the robot using AoA: 225 remaining peaks e) Proposed approach: Peak detection, filtering on peak properties, SNR-score, PDoA, calculating object position and clustering: 89 remaining peaks. The front object is not marked as this is outside the filtered \ang{-45} - \ang{45} field of view range of the side antenna.}
    \label{fig:proposed-approach-steps}
\end{figure*}

\rev{The evolution of the complete algorithm is illustrated in Fig. \ref{fig:proposed-approach-steps}. For this visualisation, data were collected using the parallel antenna on channel 5 in an environment with two objects (illustrated in Fig. \ref{fig:second-data-collection}). The filtering parameters from Table \ref{tab:parameter-results} and mapping parameters from Table \ref{tab:used-param} are used.} The mobile robot moves in a straight line from start (indicated with a semi-transparent blue circle) to end (indicated with a fully opaque blue circle). Each sub-figure includes a colour bar on the right-hand side representing the SNR-score where a brighter colour indicates higher SNR-score and thus better signal quality.

\rev{In the first step (Fig. \ref{fig:step-0}), a radargram-style visualisation illustrates the robot's path and the signal quality of the detected peaks, represented by their SNR-score. The data are plotted in an x/y plane to facilitate understanding of subsequent processing steps. In this step no angle of arrival is used, and every detected peak is shown in front of the antenna. At this point, many unwanted peaks caused by noise and unwanted multipath effects are visible, which makes raw CIR unsuitable for direct mapping since it could easily lead to false object detections. To remove the unwanted peaks, the data is filtered in the next stage (Fig. \ref{fig:step1}) using \textit{width}, \textit{prominence}, and \textit{SNR-score}}. This step significantly reduces unwanted peaks and reveals a clear pattern. It also highlights an important behaviour of the parallel antenna: although it is mounted on the right side of the robot, it still detects peaks created from objects that are not directly in front of it. In this example, peaks from an object located in front of the robot but outside the parallel antenna's forward-facing range are still recorded, resulting in a line of phantom peaks as shown in the figure diagonally down to the right.

To avoid such misleading detections, the following step (Fig. \ref{fig:step2}) introduces filtering based on the PDoA. Only peaks with AoA between \ang{-45} - \ang{45} are retained, since research has shown PDoA to be reliable within this range. Beyond these limits, phase wrapping can produce misleading estimates. Once this filtering is applied, the AoA is calculated from the phase difference (Fig. \ref{fig:step3}) and the peaks are mapped relative to the robot's location, which in this case is tracked using the MOCAP system.
The final step (Fig. \ref{fig:step4}) applies clustering to the remaining peaks and all remaining peaks within a cluster are shown. Detected components that are close to each other are grouped into clusters, which likely represent real objects. The clustering is performed after more than \textit{min\_peaks} peaks have been recorded. Peaks that do not belong to any cluster are considered noise and discarded. As a result, the remaining clusters are assigned consistent SNR-based scores, which strengthens the confidence that they correspond to actual objects in the environment.

\begin{figure}
    \centering
    \includegraphics[width=1\linewidth]{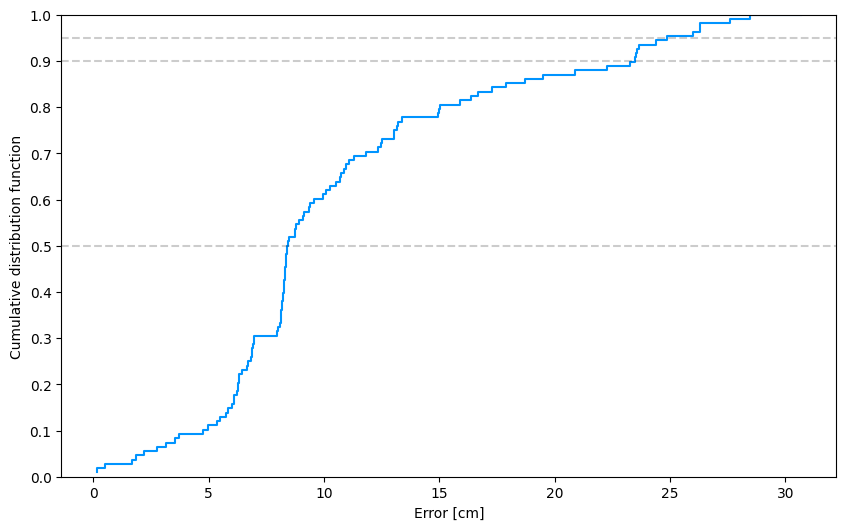}
    \caption{After step 3: CDF illustrating the errors between the estimated object position and ground truth, 90\% of the errors are below 23.50 cm and 95\% below 25.65 cm.}
    \label{fig:cdf-step4}
\end{figure}

The accuracy of the end result of the overall processing pipeline is evaluated in Fig. \ref{fig:cdf-step4}, which presents the cumulative distribution function (CDF) of the errors compared to the ground truth. The results show a median error of 8.48 cm. Moreover, 90\% of the errors are below 23.50 cm and 95\% below 25.65 cm, demonstrating that the proposed method achieves high accuracy with low error rates.

\subsubsection{Processing latency}
Next, this section examines the processing overhead of the full approach. The proposed pipeline achieves an average processing time of 0.48 ms per sample, and 4.73 ms when clustering is applied after 50 samples. The average processing latency is measured on a Windows 11 laptop with an Intel(R) Core(TM) Ultra 7 165U CPU and 16 GB RAM. Both values remain below the 10.42 ms interval between two samples at 96 Hz (Table \ref{tab:uwb-specifications}).

\subsubsection{Visual Demonstration of the Proposed Approach}
\begin{figure}
    \centering
    \includegraphics[width=0.75\linewidth]{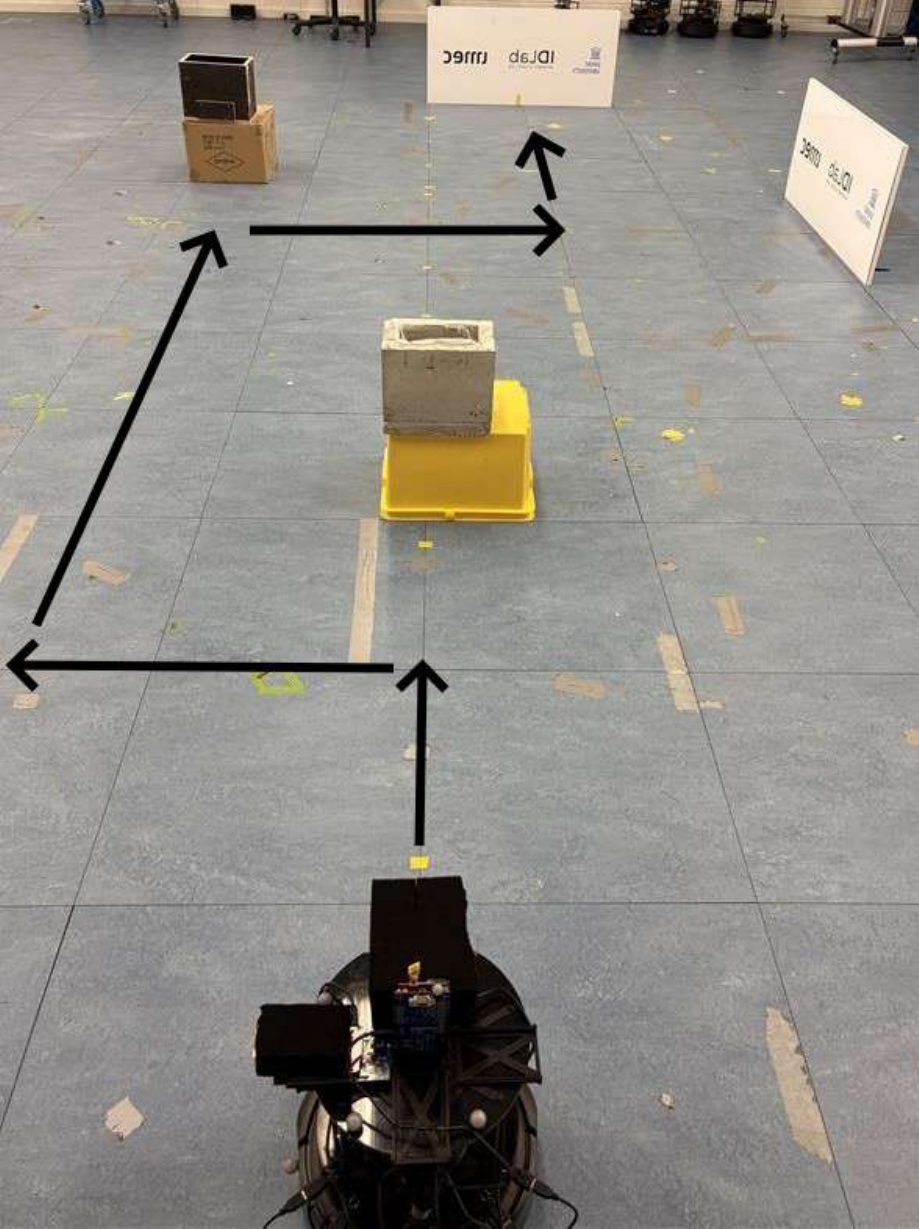}
    \caption{The lab setup containing four obstacles, the robot and its followed path.}
    \label{fig:demo-lab}
\end{figure}

\begin{figure}
    \centering
    \includegraphics[width=0.75\linewidth]{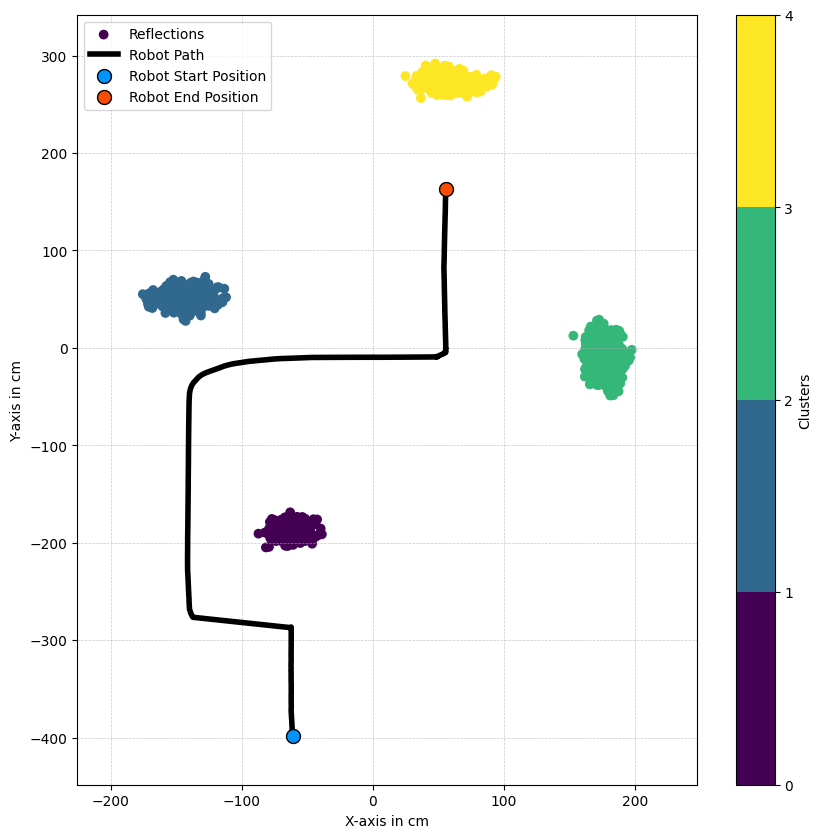}
    \caption{The resulting map showing the detected obstacles and their localisation using our proposed approach.}
    \label{fig:end-demo}
\end{figure}

From the experimental results, it can be concluded that channel 9 achieves the highest performances across all three tested material types. To provide a clearer understanding of this outcome, a visual demonstration is made available online\footnote{https://www.youtube.com/watch?v=maDkhShUsjw}.
The algorithm is not executed in real time in this instance, instead, the recorded data is processed offline and subsequently aligned with the recorded video.
Fig. \ref{fig:demo-lab} illustrates the experimental setup, which consists of four obstacles (metal, concrete, and plywood), the mobile robot, and its followed path. Fig. \ref{fig:end-demo} presents the resulting environment map generated using our proposed approach.

\section{Discussion}
\label{sec:discussion}
\rev{This research identified two main limitations. First, objects closer than approximately 60 cm are not detected. In this near-field region, the reflection peak merges with the strong direct path between transmitter and receiver, making separation impossible. This effect is primarily caused by insufficient isolation between the antennas, but material properties further influence detectability. Highly reflective objects still generate visible peaks, while absorptive materials are easily masked. Improved antenna isolation between the transmitter and receiver could help reduce the direct-path signal, or higher bandwidth settings could be used to achieve better timing resolution. Although this blind zone prevents near-field detection, objects are consistently detected at larger distances, allowing obstacle avoidance or SLAM systems to react before entering the 60 cm range. For tasks requiring the robot to operate closer, this limitation can be handled by fusing UWB radar data with a close-range sensor.}
\rev{Second, the analysis showed that peak prominence decreases and width increases with distance, however, the filters used in this work employed fixed thresholds.}

\section{Conclusion}
\label{sec:conclusion}
\rev{This paper presented an infrastructure-free UWB radar-based approach for obstacle mapping on mobile robots. The proposed method consists of three steps: (i) target identification (based on CIR peak detection), (ii) filtering (based on peak properties, SNR-based score, and phase-difference of arrival), and (iii) clustering (based on distance estimation and angle-of-arrival estimation). The experimental evaluation across channels 5 and 9 and three material types demonstrated that the proposed system achieved high distance-estimation accuracy and detection accuracy, with obstacle localisation errors below 17 cm. Even in challenging low-reflectivity scenarios such as concrete, the method achieves a precision of 73.42\% and a recall of 83.38\% on channel 9.}
\rev{The approach enables lightweight, anchor-free mapping and provides a solid foundation for future UWB-based SLAM and other robotic applications. Remaining challenges include detection at close range (< 60 cm) and improving antenna isolation to reduce direct-path interference. Future work may explore adaptive thresholds for peak properties, improved antenna designs, or fusion with complementary short-range sensing to address these limitations.}

\bibliography{references.bib}

\phantomsection
\begin{IEEEbiography}[{\includegraphics[width=1in,height=1.25in,clip,keepaspectratio]{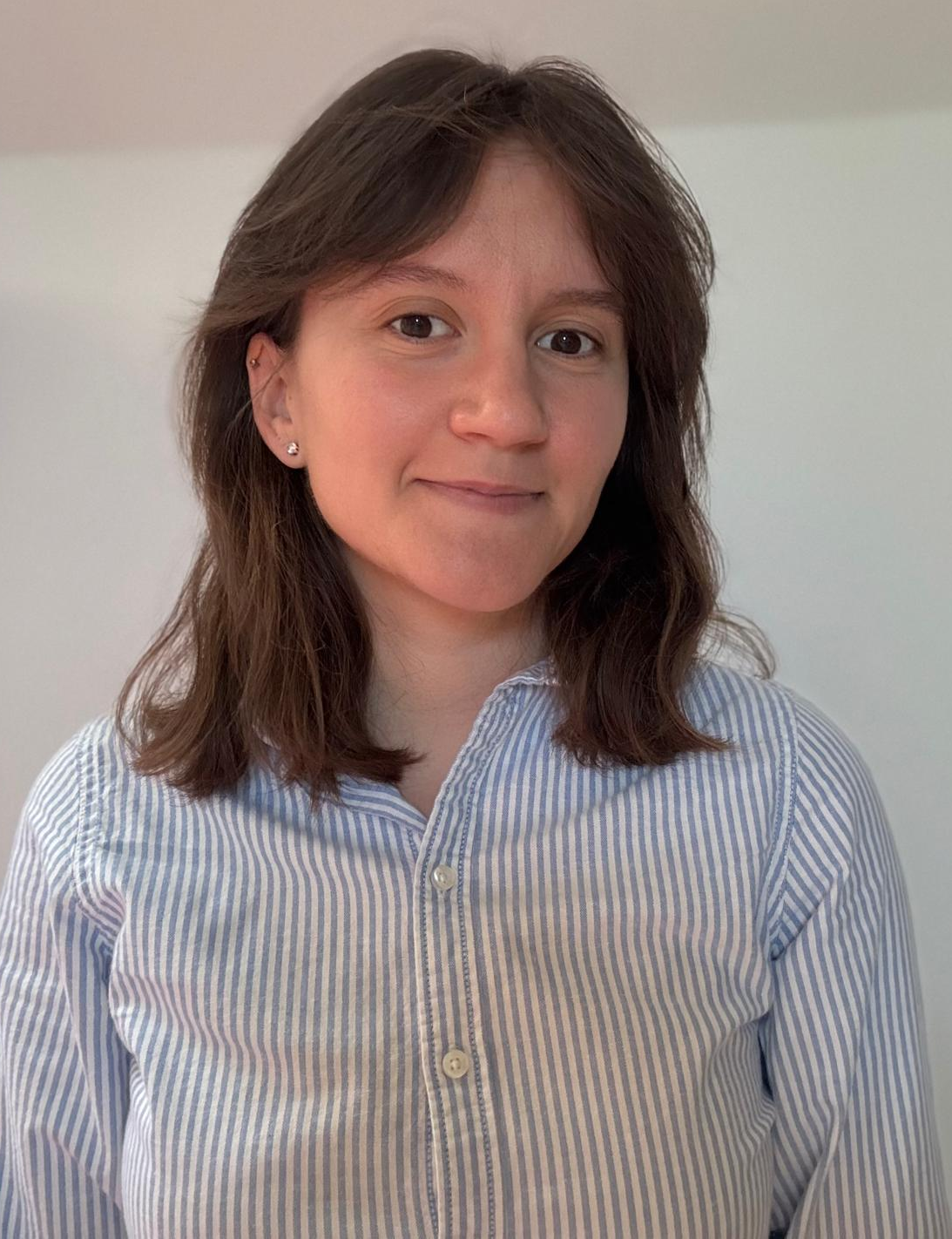}}]{Adelina Giurea} received the M.Sc. degree in Information Engineering Technology from Ghent University, Belgium, in September 2025. She is currently pursuing the Ph.D. degree with the IDLab Research Group, focusing on UWB and BLE.
\end{IEEEbiography}

\begin{IEEEbiography}[{\includegraphics[width=1in,height=1.25in,clip,keepaspectratio]{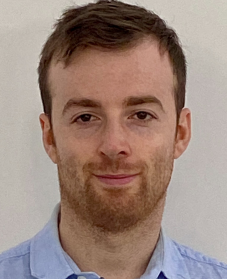}}]{Ir. Stijn Luchie} was born in Veurne in 1999. He received his M.Sc. degree in Computer Science Engineering from Ghent University, Belgium, in September 2022. Shortly thereafter, in October 2022, he started as a researcher with the Department of Information Technology (INTEC) at Ghent University in the IDLab research group. His primary area of scientific interest centres on ultra-wideband technology, with a specific focus on the self-calibration of anchor networks and the applications of UWB radar.
\end{IEEEbiography}

\begin{IEEEbiography}[{\includegraphics[width=1in,height=1.25in,clip,keepaspectratio]{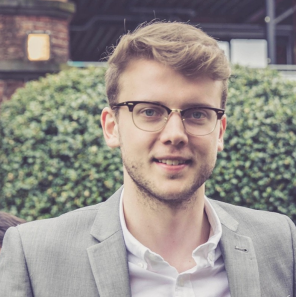}}]{Dr. ir. Dieter Coppens} received the master’s degree in electrical engineering from Ghent University, Belgium, in 2021 and the Ph.D. degree with the IDLab Research Group in 2025. His research interests are wireless networking, indoor localisation systems based on Ultra-wideband technology and machine learning.
\end{IEEEbiography}

\begin{IEEEbiography}[{\includegraphics[width=1in,height=1.25in,clip,keepaspectratio]{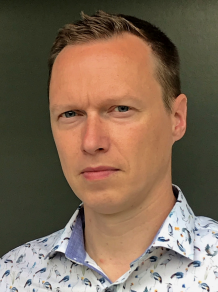}}]{Prof. dr. ir. Jeroen Hoebeke} received the Master's degree in Engineering Computer Science from Ghent University in 2002. In 2007, he obtained a Ph.D. in Engineering Computer Science with his research on adaptive ad hoc routing and Virtual Private Ad Hoc Networks. Currently, he is an associate professor in the Internet Technology and Data Science Lab of Ghent University and imec. He is conducting and coordinating research on wireless (IoT) connectivity, embedded communication stacks, deterministic wireless communication and wireless network management. He is author or co-author of more than 200 publications in international journals or conference proceedings.
\end{IEEEbiography}

\begin{IEEEbiography}[{\includegraphics[width=1in,height=1.25in,clip,keepaspectratio]{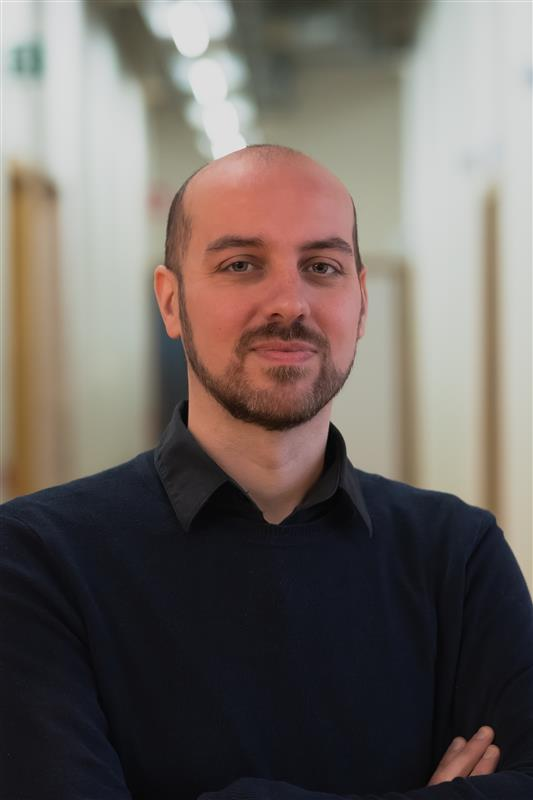}}]{Prof. dr. ir. Eli De Poorter} is currently a Professor with the IDLab Research Group, Ghent University, and imec. His team performs research on wireless communication technologies such as (indoor) localisation solutions, wireless IoT solutions, and machine learning for wireless systems. He performs both fundamental and applied research. For his fundamental research, he is currently the coordinator of several research projects (SBO, FWO, and GOA). He has over 200 publications in international journals or in the proceedings of international conferences. For his applied research, he collaborates with industry partners to transfer research results to industrial applications, and to solve challenging industrial research problems.
\end{IEEEbiography}

\EOD

\end{document}